%% file: main.tex
\documentclass[letterpaper]{article} \usepackage{aaai2027} \nocopyright  \usepackage[hyphens]{url}  \usepackage{graphicx}    \usepackage{natbib}  \usepackage{caption}   \usepackage{amsmath}
\usepackage{multirow}

\usepackage{booktabs}

\usepackage[table]{xcolor}
\usepackage{algorithm}
\usepackage{algorithmic}
\usepackage{tcolorbox}
\tcbuselibrary{breakable,skins}
\usepackage{fontawesome5}
\usepackage{colortbl}
\usepackage{tabularx}
\usepackage{arydshln}
\usepackage{pifont}
\usepackage{array}
\usepackage{ragged2e}
\newcommand{\cmark}{\ding{51}}
\newcommand{\xmark}{\ding{55}}
\newcolumntype{P}[1]{>{\RaggedRight\arraybackslash}p{#1}}
\newcommand{\projectpage}{\pdfstartlink attr{/Border [0 0 0]} user{/Subtype /Link /A << /S /URI /URI (https://theappliedscientist.github.io) >>}\url{https://theappliedscientist.github.io}\pdfendlink}
\DeclareUnicodeCharacter{00D7}{\ensuremath{\times}}
\DeclareUnicodeCharacter{2265}{\ensuremath{\geq}}

\definecolor{PastelGreen}{RGB}{226,239,218}
\definecolor{PastelRed}{RGB}{252,228,214}
\definecolor{excluded}{HTML}{F8F8F8}
\definecolor{included}{HTML}{EEF2FF}

\title{AppliedScientist: Automated Scientific Revision Through Iterative AI Reviewing}
\author{
    Vidushee Vats\equalcontrib,
    Karun Sharma\equalcontrib,
    Shengzhi Li\equalcontrib,
    Shichao Pei\textsuperscript{\rm 1}
}
\affiliations{
    \textsuperscript{\rm 1}Department of Computer Science\\
    University of Massachusetts Boston
}

\begin{document}

\maketitle

\begin{abstract}
Automated reviewing systems are increasingly evaluated based on the quality of the reviews they produce. Yet a review is only useful if acting on it leads to a measurable improvement in the paper. We present AppliedScientist, a closed-loop system that couples an autonomous AI scientist with an AI reviewer, and evaluate it by iteratively revising rejected papers from a range of research subfields. To mirror how human authors build on earlier drafts, the AI scientist has access to its previous versions during revision. To avoid bias from prior judgments, however, each review is generated independently, with the reviewer having no memory of earlier feedback or scores. We compare three revision settings: one initialized with the original venue reviews, one initialized with AI-generated reviews, and autonomous self-revision using the same fixed prompt in every round. Because the reviewer both guides and evaluates the revision, we also assess the human-initialized revisions using Stanford Reviewer as an independent evaluator. Reviewer-guided revision consistently improves more than fixed-prompt self-revision, and Stanford Reviewer also assigns higher scores to later revisions. AppliedScientist resolves 128 of 150 execution-related weaknesses (85.3\%), but only 2 of 18 idea-related weaknesses (11.1\%), suggesting that iterative revision is effective at improving experiments and implementation, but rarely changes concerns about novelty or significance.

\noindent\textbf{Project page:} \projectpage

\end{abstract}

\section{Introduction}
Scientific publishing is an iterative process rather than a one-time evaluation. Reviewers identify strengths and weaknesses, and authors revise the work before it is considered again. Addressing reviewer feedback, however, often requires substantially more than revising the manuscript. Authors may need to reproduce results, run new experiments, evaluate stronger baselines, and determine whether the resulting evidence addresses the reviewers' concerns. As language models become increasingly capable of conducting research and reviewing scientific papers, a natural question emerges: Can this process of scientific review and revision itself be automated?

Recent work has steadily automated different stages of the scientific publication process. AI scientists can conduct research and prepare manuscripts \citep{AIS, AIS2, co-scientist}, while AI reviewers can produce detailed critiques grounded in a submitted paper \citep{deepreview, scholarpeer, stanford}. Yet generating a paper or a review does not establish that acting on the review improves the underlying research. In a closed revision loop, the reviewer is not merely an evaluator; it supplies the feedback signal that determines which weaknesses the scientist addresses, much as a grader or reward model directs optimization in reinforcement learning. Even a capable scientist can therefore optimize toward the wrong objective when its reviewer is unreliable. Existing systems provide either a broad experimental action space under fixed human feedback or dynamic AI feedback restricted to paper generation or textual rewriting. What remains unclear is whether an independently evaluated AI reviewer can repeatedly guide code, experiment, and manuscript revisions of the same existing paper without replacing its central contribution.

We study this question using real rejected papers, which already contain an implemented method, experimental code, and reviewer feedback describing the work's shortcomings. We introduce \textbf{AppliedScientist}, a closed-loop system in which a scientist revises the implementation, executes experiments, analyzes the resulting artifacts, and updates the manuscript. After each revision, an independent \textbf{AI Reviewer} evaluates only the current manuscript and returns fresh feedback for the next round. The scientist retains its previous code, results, manuscripts, and feedback so that revisions accumulate and the reviewer retains no history, preventing earlier judgments or scores from biasing its assessment of the current version. The central research contribution remains fixed, allowing us to study revision rather than replacement of the original project.

We evaluate three sources of revision guidance. In the \emph{human-initialized} condition, written venue reviews guide the first round and the AI Reviewer guides subsequent rounds. In the \emph{AI-initialized} condition, the AI Reviewer also supplies the first-round feedback. In \emph{autonomous self-revision}, the scientist instead receives the same fixed self-review prompt in every round. The scientist has the same research capabilities in all three conditions, isolating the effect of the guidance it receives. Because our AI Reviewer both guides and scores the reviewer-driven conditions, we additionally use Stanford Reviewer to evaluate every saved version from the human-initialized trajectories.

We evaluate AppliedScientist on 25 rejected and 5 borderline-accepted ICLR papers spanning multiple research domains. Reviewer-guided revision improves substantially more than autonomous self-revision, while Stanford Reviewer also assigns higher scores to later human-initialized versions. The effect depends strongly on the type of weakness: AppliedScientist resolves 128 of 150 execution-related weaknesses (85.3\%), but only 2 of 18 idea-related weaknesses ($\sim 11.1\%$). Iterative revision can therefore strengthen the implementation, experimental evidence, and presentation of an existing contribution, but it rarely resolves concerns about the novelty or significance of the contribution itself

Our contributions are three-fold:
\begin{itemize}
    \item We introduce AppliedScientist, an iterative scientist-reviewer system that revises an existing paper's implementation, experiments, and manuscript in response to fresh reviews of each version.

    \item We treat the reviewer as a first-class component: we benchmark its review quality separately, compare its dynamic guidance with fixed-prompt self-revision, and use an external reviewer to evaluate the resulting trajectories.

    \item We characterize what autonomous revision can and cannot improve: execution-related concerns can often be resolved through systematic revision, whereas novelty-related concerns remain substantially more difficult.
\end{itemize}

\section{Related Work}

\paragraph{AI Scientists and Autonomous Research} Autonomous AI systems increasingly cover multiple stages of the scientific research lifecycle, including idea generation, literature review, experimentation, manuscript preparation, and peer review \citep{AIS, AIS2, co-scientist, research-agent, cycle, robin, kosmos, agentlab}. Prior systems already recognize the value of reviewing, but use it in different ways. AI Scientist evaluates a completed manuscript and archives the result for future research generations, AI Scientist-v2 uses VLM critiques to refine experimental and manuscript figures, and CycleResearcher uses CycleReviewer scores to construct preference pairs across training rounds \citep{AIS, AIS2, cycle}. Robin and Kosmos push the same loop into new domains: Robin runs a full hypothesis-generate-and-test cycle for experimental biology and Kosmos shares a structured model across parallel analysis and literature-search agents to sustain longer, coherent discovery sessions \citep{robin, kosmos}. These works establish strong foundations for autonomous scientific research. In contrast, our work focuses on iterative scientific revision guided by repeated peer review.

\paragraph{Automated Peer Review} Automated reviewers increasingly use structured reasoning, multiple agents, and literature retrieval to produce grounded critiques \citep{deepreview, scholarpeer, stanford, openreview, remor, deepreviewev2}. These systems differ mainly in how they get to a grounded critique: OpenReviewer fine-tunes directly on expert reviews, Remor trains with reinforcement learning against a reward function built from human preferences, and DeepReviewer 2.0 forces every claim in its output to be traceable to a specific location in the manuscript before it is released \citep{openreview, remor, deepreviewev2}.\citet{scholarpeer}, \citet{agenrreview}, \citet{marg}, \citet{stanford} and \citet{multiturn} focus on the use of multiple agents, distinct personas, and retrieval mechanisms to simulate collaborative reviewing.  This literature primarily evaluates the review itself. We separately benchmark our reviewer, and then observe whether acting on its feedback improves the research.

\paragraph{AI-Assisted Scientific Revision} Compared with automated reviewing, autonomous scientific revision remains relatively underexplored. The closest systems each cover one side of our setting. Apres revises existing scientific manuscripts by optimizing a learned rubric predictive of future citation impact, but restricts its action space to presentation and protects experimental results from modification \citep{apres}. CycleResearcher studies the author-reviewer interaction during automated paper generation, but operates on synthetically generated manuscripts and does not execute new experiments during refinement \citep{cycle}. Alongside these systems, recent resources have emerged for studying revision itself. \citet{revisearena} benchmark an LLM's ability to revise papers using real reviewer feedback, allowing repository edits, experiment execution, and manuscript revision, but acts once on a fixed set of human weaknesses rather than obtaining a new review of the revised paper. Other systems stay at the level of text entirely. ARIES aligns reviewer comments to the edits they prompted in the original manuscript, without generating new edits itself, and XtraGPT, trained on 140K instruction–revision pairs derived from published papers, rewrites one section at a time (e.g. the introduction) to satisfy a stated instruction, without touching the underlying experiments \citep{aries, xtra}. Together, these works establish revision as an emerging research direction, although they primarily focus on textual revision rather than experimentally validating reviewer feedback.

\section{Methodology}

\begin{figure*}[t]
    \centering
    \begin{picture}(505,209)
        \put(0,0){\includegraphics[width=\textwidth]{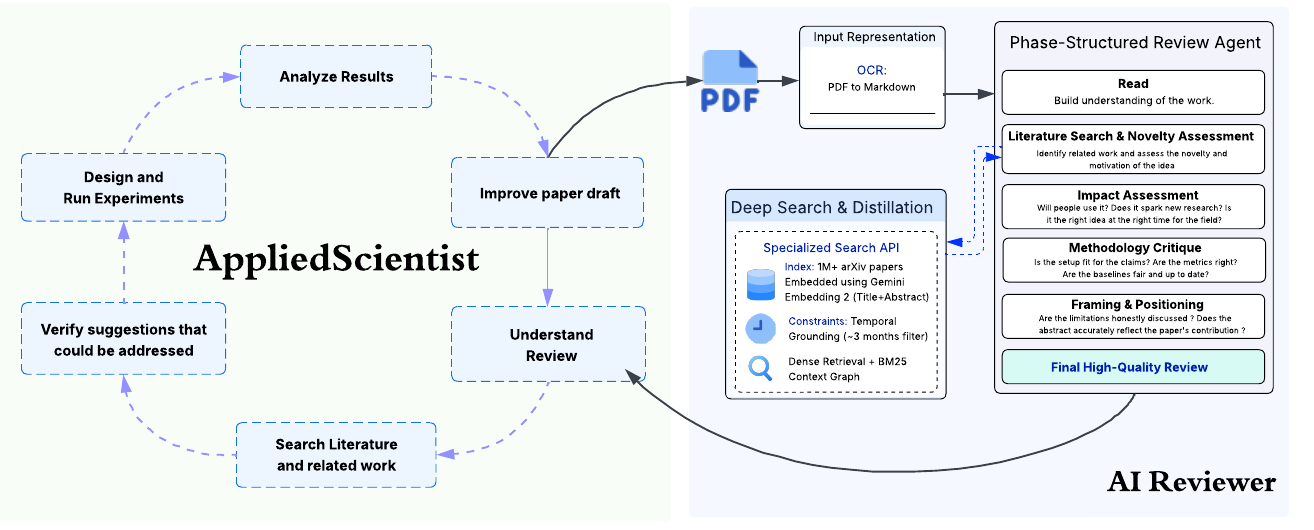}}
        \put(375,18){\colorbox{white}{\parbox[c][24pt][c]{125pt}{\centering\Large\bfseries AI Reviewer}}}
    \end{picture}
    \caption{Overview of AppliedScientist. The scientist iteratively verifies reviewer suggestions, runs experiments, analyzes results, and updates the manuscript. The review component independently reads each revision, searches the literature, assesses its technical quality and significance, and returns structured feedback for the next round.}
    \label{fig:system_overview}
\end{figure*}

AppliedScientist consists of two components: a scientist that revises the implementation, experiments, and manuscript, and an AI reviewer that independently reviews each revision and provides feedback for the next iteration. We describe each component below.

\paragraph{AppliedScientist}
Revising an existing scientific paper differs fundamentally from generating one from a new research idea. During research generation, the problem formulation, methodology, and experimental design may all evolve as the work develops. Revision instead begins with an established contribution, implementation, and manuscript, where the goal is to address reviewer concerns without replacing the central research idea. AppliedScientist is designed for this constrained setting.

Given a rejected paper, its source repository, and current feedback, the scientist is instructed to address every reviewer concern while determining how each should be resolved. Depending on the feedback, this may require inspecting the repository and manuscript, searching the literature, editing implementation, reproducing results, adding baselines and presenting ablations, executing new experiments and analyzing the results and updating the manuscript. The central research contribution is treated as fixed; if addressing a concern would require changing that contribution, the scientist records it as unresolved rather than reframing the work as a different project.

Throughout the revision process, the scientist maintains the complete revision history, including previous manuscript versions, reviewer feedback, experimental results, and intermediate analyses. Each revision builds upon the results and decisions from earlier iterations rather than restarting from the original submission. The revision procedure is otherwise identical across all three conditions; only the guidance provided to the scientist changes.

\paragraph{Review Component}

Our reviewer is inspired by how human reviewers evaluate scientific papers. Rather than assessing every aspect of a paper simultaneously, reviewers gradually build an understanding of the work before forming specific judgments. For example, novelty cannot be assessed without understanding the relevant literature, and experimental results cannot be evaluated without first understanding the claims they are intended to support. Skipping or reordering these steps can lead to unreliable assessments. Our reviewer therefore follows a structured review process in which each stage builds upon the information established by the previous one.

The reviewer first identifies the paper's contributions, methodology, and evaluation setting before conducting an independent literature search beyond the cited references. To assess novelty, it performs a deep search over an indexed collection of around one million arXiv papers. The search retrieves literature from multiple perspectives, including direct competitors, alternative approaches to the same problem, recent benchmark methods, and work built upon similar baselines. It then evaluates the paper's novelty, positioning with respect to prior work, the adequacy of its experimental evaluation, and its broader impact before synthesizing these observations into a structured review containing strengths, weaknesses, and suggestions for the next revision.

Unlike the scientist, our reviewer intentionally maintains no memory of previous revisions. Every manuscript version is reviewed independently without access to earlier drafts, reviewer comments, or historical scores. Retaining this history could bias the review by allowing judgments to be influenced by previous evaluations or previous scores rather than the quality of the submitted manuscript. Before producing the final review, the reviewer performs a verification pass to ensure that factual statements and comparisons are supported by either the manuscript or the retrieved literature.

\paragraph{Revision Loop}
Let $V_0$ and $C_0$ denote the original manuscript and codebase. At revision round $t$, the scientist receives the previous manuscript and codebase $(V_{t-1}, C_{t-1})$, its accumulated revision history, and guidance $G_{t-1}$. It then updates the implementation, performs any required experiments, and produces revised code, experimental artifacts, and manuscript $(C_t, E_t, V_t)$. The reviewer evaluates only $V_t$ (together with any literature retrieved during that review) and generates a new review $R_t$ without access to earlier manuscript versions, reviews, or evaluation scores. In the reviewer-guided conditions, $G_t = R_t$ for the next revision round. In the autonomous self-revision condition, $G_t$ remains the same fixed self-review prompt, while $R_t$ is used only for evaluation.

\begin{figure*}[t]
    \centering
    \includegraphics[width=\textwidth]{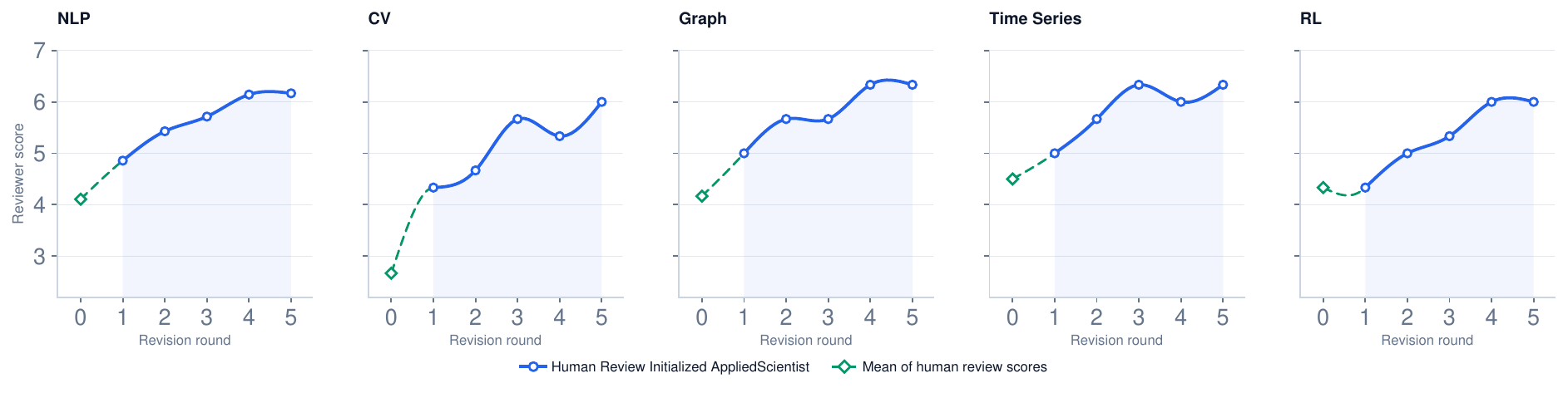}
    \caption{Score trajectories across five scientific domains for the human-initialized condition. Round 0 reports the mean score assigned to the original manuscript by the human venue reviewers. AppliedScientist uses the written venue reviews to produce Round 1. Thereafter, our AI reviewer independently reviews each revision, and its feedback guides the following round.}
    \label{fig:domain}
\end{figure*}

\section{Experimental Setup}

\paragraph{Data} We mine 25 rejected and 5 borderline-accepted ICLR papers with both publicly available reviews and code. To ensure a diverse evaluation set, we sample papers spanning multiple research domains while preferring those with low variance in original human reviewer scores (Table~\ref{tab:benchmark}). Finally, we filter papers by computational requirements, retaining only those that can be executed on CPU or a single GPU with 96 GB of VRAM.

\begin{table}[t]
\centering
\small
\begin{tabular}{lccc}
\toprule
& & \multicolumn{2}{c}{\textbf{Decision}} \\
\cmidrule(lr){3-4}
\textbf{Domain} & \textbf{N} & \textbf{Rejected} & \textbf{Borderline} \\
\midrule
NLP                    & 12 & 11 & 1 \\
Computer Vision        & 4  & 3  & 1 \\
Graph Learning         & 5  & 4  & 1 \\
Time Series            & 5  & 4  & 1 \\
Reinforcement Learning & 4  & 3  & 1 \\
\midrule
\textbf{Total} & \textbf{30} & \textbf{25} & \textbf{5} \\
\bottomrule
\end{tabular}
\caption{We curate 30 ICLR papers spanning five research domains and retain only papers with publicly available source code that can be reproduced on a single GPU within our VRAM budget.}
\label{tab:benchmark}
\end{table}

\paragraph{Setup} We evaluate AppliedScientist under the three revision conditions summarized in Table~\ref{tab:revision_conditions}. The human-initialized condition begins with the original venue reviews, whereas the AI-initialized condition replaces them with reviews generated by our reviewer. In both cases, our reviewer evaluates every subsequent revision, and each new review guides the next iteration. By contrast, the autonomous self-revision condition uses no external reviewer feedback and the scientist relies on a same self-review prompt throughout. This comparison isolates the effect of iterative reviewer feedback from the scientist's ability to revise autonomously. For all conditions, $V_0$ denotes the original manuscript and $V_1$--$V_5$ denote successive revisions. Each revision run takes approximately 9 hours on a 96 GB VRAM GPU.

\begin{table*}[t]
\centering
\small
\setlength{\tabcolsep}{6pt}
\begin{tabular}{p{0.20\textwidth}p{0.25\textwidth}p{0.25\textwidth}p{0.20\textwidth}}
\toprule
\textbf{Condition} & \textbf{Guidance used for $V_1$} & \textbf{Guidance used for $V_2$--$V_5$} & \textbf{Role of our reviewer} \\
\midrule
Human-initialized & Original written venue reviews of $V_0$ & A fresh, independent AI review of the preceding version & Guides $V_2$--$V_5$ and scores each version \\
AI-initialized & A fresh, independent AI review of $V_0$ & A fresh, independent AI review of the preceding version & Guides $V_1$--$V_5$ and scores each version \\
Autonomous self-revision & Fixed self-review prompt & The same fixed self-review prompt in every round & Evaluation only; no feedback is shown to the scientist \\
\bottomrule
\end{tabular}
\caption{Revision conditions. Guidance is the feedback shown to the scientist before it produces the indicated version. Evaluation-only reviews and scores are never exposed to the scientist. Stanford Reviewer is applied only to the saved versions from the human-initialized condition.}
\label{tab:revision_conditions}
\end{table*}

\paragraph{Evaluation}
Our reviewer scores every saved manuscript version without access to earlier drafts, feedback, or scores. Because this reviewer also guides the reviewer-driven conditions, we separately score the human-initialized trajectory with Stanford Reviewer. Stanford Reviewer is used only for external evaluation and its outputs are never shown to the scientist. The score and threshold analyses below use the 30 completed human-initialized trajectories.

\section{Reviewer Evaluation}
In a closed revision loop, the reviewer serves both as an evaluator and as the feedback signal that guides the scientist's next action. Reviewer validity is therefore a prerequisite for interpreting downstream improvements: a stronger scientist does not make the loop reliable if its reviewer consistently rewards the wrong changes. Before using our reviewer's scores to assess revision, we compare it with existing automated reviewers using the evaluation setup introduced in DeepReviewBench and ScholarPeer \citep{deepreview,scholarpeer}. We evaluate on 300 papers from 2020--2026, using the ICLR 2024 and ICLR 2025 splits of DeepReviewBench together with 300 papers from the AgentReviewer test set containing ICLR papers from 2020--2023. To evaluate performance on newer research, we additionally include a held-out set of 50 ICLR 2026 submissions. We adopt the extensions to the original DeepReviewBench protocol, replace the plausibility criterion with significance assessment, and allow the judge to use web search for novelty verification, following ScholarPeer.

We use a search-enabled judge that receives the input paper together with reviews from two anonymized systems, with review order randomized to avoid positional bias. The judge compares the reviews across five dimensions: Technical Accuracy, Constructive Value, Analytical Depth and Significance Assessment, with strong penalties for hallucinated claims and generic feedback. Table~\ref{tab:judge_comparison} shows that our Reviewer outperforms all baselines across nearly every evaluation dimension. On Constructive Value, however, it achieves performance comparable to Stanford Agent Reviewer (44.3\% vs. 45.7\% win rate), with the remaining 10\% of comparisons resulting in ties.

\begin{figure}[t]
    \centering
    \includegraphics[width=\columnwidth]{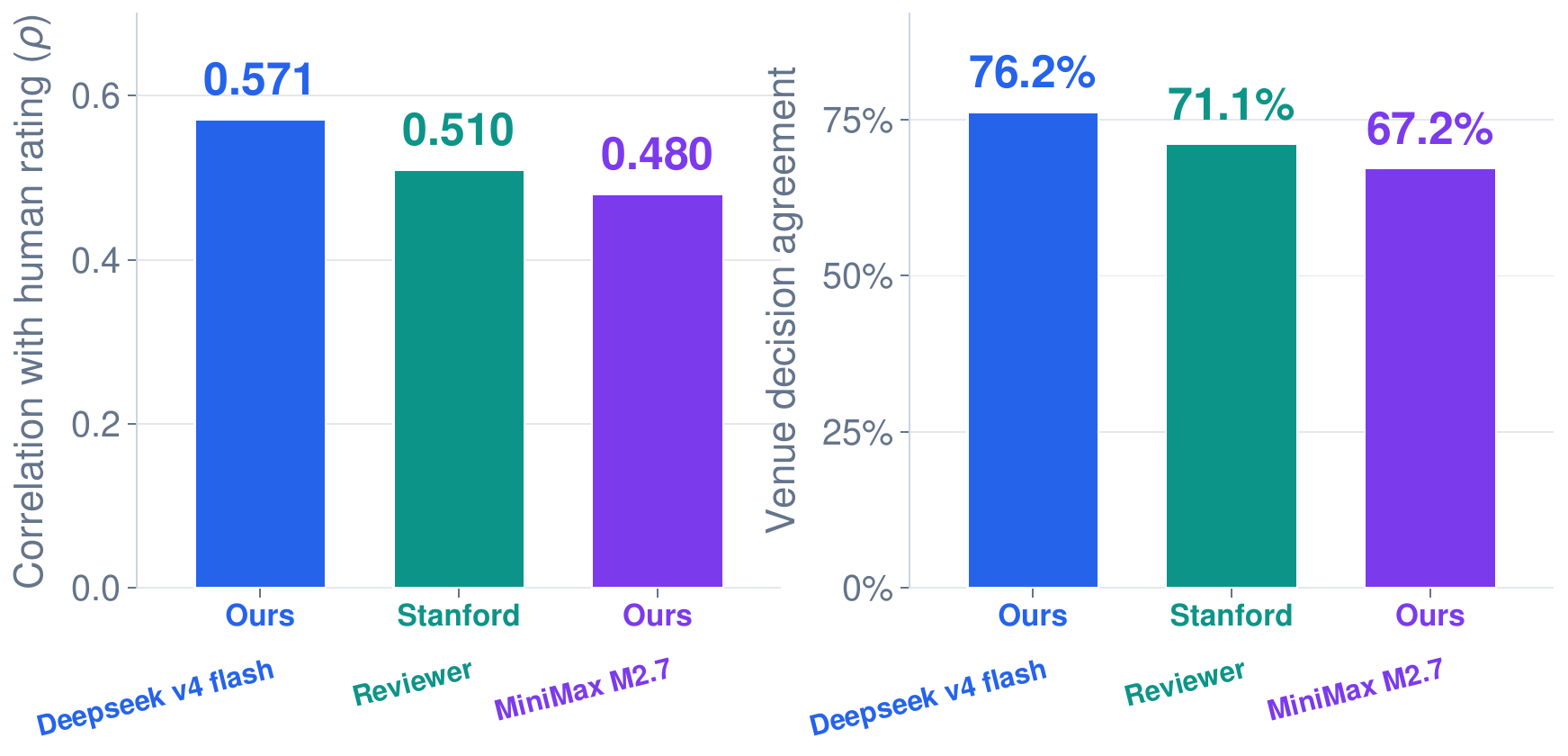}
    \caption{Score alignment with venue reviews on a set of 650 ICLR 2020-26 papers. Left: Spearman correlation with the mean human rating. Right: agreement
    with the venue's accept/reject decision.}
    \label{fig:alignment}
\end{figure}

Across these papers, our Reviewer 
correlates with mean human ratings at $\rho = 0.571$ with DeepSeek V4 Flash and matches the venue's decision $76.2\%$ of the time, against $\rho = 0.48$ and $67.2\%$ with MiniMax M2.7 (Figure~\ref{fig:alignment}). The Stanford Reviewer falls between the two, at $\rho = 0.51$ and $71.1\%$.

\begin{table*}[t]
\small
\setlength{\tabcolsep}{2pt}      \renewcommand{\arraystretch}{1.05}
\begin{tabular}{llcccccccc}
\toprule

\multirow{2}{*}{\textbf{Category}} &
\multirow{2}{*}{\textbf{Baseline}} &
\multicolumn{2}{c}{\shortstack{\textbf{Technical}\\\textbf{Accuracy}}} &
\multicolumn{2}{c}{\shortstack{\textbf{Constructive}\\\textbf{Value}}} &
\multicolumn{2}{c}{\shortstack{\textbf{Analytical}\\\textbf{Depth}}} &
\multicolumn{2}{c}{\shortstack{\textbf{Significance}\\\textbf{Assessment}}} \\

\cmidrule(lr){3-4}
\cmidrule(lr){5-6}
\cmidrule(lr){7-8}
\cmidrule(lr){9-10}

&
& \textbf{Win (\%)} & \textbf{Loss (\%)}
& \textbf{Win (\%)} & \textbf{Loss (\%)}
& \textbf{Win (\%)} & \textbf{Loss (\%)}
& \textbf{Win (\%)} & \textbf{Loss (\%)} \\

\midrule

\multirow{4}{*}{\textbf{Fine-tuned}}

& CycleReviewer-8B
& \cellcolor{PastelGreen}100 & 0
& \cellcolor{PastelGreen}100 & 0
& \cellcolor{PastelGreen}99 & 1
& \cellcolor{PastelGreen}100 & 0 \\

& CycleReviewer-70B
& \cellcolor{PastelGreen}99.2 & 0.4
& \cellcolor{PastelGreen}100 & 0
& \cellcolor{PastelGreen}100 & 0
& \cellcolor{PastelGreen}99.6 & 0.2 \\

& DeepReviewer-7B
& \cellcolor{PastelGreen}96.5 & 2.1
& \cellcolor{PastelGreen}99.4 & 0.4
& \cellcolor{PastelGreen}99.2 & 0.6
& \cellcolor{PastelGreen}97.8 & 1.1 \\

& DeepReviewer-14B
& \cellcolor{PastelGreen}91.2 & 3.7
& \cellcolor{PastelGreen}96.0 & 2.2
& \cellcolor{PastelGreen}94.7 & 2.6
& \cellcolor{PastelGreen}97.9 & 1.3 \\

\midrule

\multirow{4}{*}{\textbf{LLM}}

& Gemini 3.1 Flash
& \cellcolor{PastelGreen}60.6 & 21.5
& \cellcolor{PastelGreen}88.2 & 6.6
& \cellcolor{PastelGreen}75.1 & 9.9
& \cellcolor{PastelGreen}74.2 & 4.6 \\

& Gemini 3.1 Pro
& \cellcolor{PastelGreen}62.2 & 11.7
& \cellcolor{PastelGreen}83.9 & 6.5
& \cellcolor{PastelGreen}71.6 & 9.0
& \cellcolor{PastelGreen}70.7 & 5.1 \\

& DeepSeek V4 Flash
& \cellcolor{PastelGreen}74.3 & 10.9
& \cellcolor{PastelGreen}92.5 & 7.1
& \cellcolor{PastelGreen}78.9 & 7.4
& \cellcolor{PastelGreen}82.3 & 3.9 \\

& DeepSeek V4 Pro
& \cellcolor{PastelGreen}78.6 & 7.5
& \cellcolor{PastelGreen}96.5 & 3.0
& \cellcolor{PastelGreen}79.2 & 5.1
& \cellcolor{PastelGreen}80.4 & 3.2 \\

\midrule
\multirow{4}{*}{\shortstack[l]{\textbf{LLM}\\\textbf{w/ Search}}}

& Gemini 3.1 Flash
& \cellcolor{PastelGreen}55.8 & 12.8
& \cellcolor{PastelGreen}85.7 & 8.5
& \cellcolor{PastelGreen}70.5 & 10.2
& \cellcolor{PastelGreen}72.8 & 6.3 \\

& Gemini 3.1 Pro
& \cellcolor{PastelGreen}56.3 & 14.6
& \cellcolor{PastelGreen}80.1 & 7.6
& \cellcolor{PastelGreen}68.4 & 8.8
& \cellcolor{PastelGreen}66.9 & 5.8 \\

& DeepSeek V4 Flash
& \cellcolor{PastelGreen}71.4 & 15.3
& \cellcolor{PastelGreen}89.4 & 7.0
& \cellcolor{PastelGreen}76.9 & 7.5
& \cellcolor{PastelGreen}80.1 & 5.0 \\

& DeepSeek V4 Pro
& \cellcolor{PastelGreen}72.9 & 17.6
& \cellcolor{PastelGreen}93.1 & 6.5
& \cellcolor{PastelGreen}73.8 & 8.1
& \cellcolor{PastelGreen}79.7 & 6.4 \\

\midrule

\multirow{6}{*}{\shortstack[l]{\textbf{Multi-Agent}\\\textbf{Framework}\\\textbf{w/ Search}}}

& Agent Review (Gemini 3.1 Pro)
& \cellcolor{PastelGreen}52.1 & 18.9
& \cellcolor{PastelGreen}89.5 & 10.4
& \cellcolor{PastelGreen}60.7 & 10.8
& \cellcolor{PastelGreen}65.6 & 7.9 \\

& Agent Review (DeepSeek V4 Flash)
& \cellcolor{PastelGreen}61.7 & 20.5
& \cellcolor{PastelGreen}92.3 & 4.0
& \cellcolor{PastelGreen}72.8 & 11.6
& \cellcolor{PastelGreen}68.5 & 8.7 \\

& AI Scientist v2 (Gemini 3.1 Pro)
& \cellcolor{PastelGreen}50.2 & 22.3
& \cellcolor{PastelGreen}84.5 & 12.1
& \cellcolor{PastelGreen}58.5 & 12.5
& \cellcolor{PastelGreen}62.2 & 9.5 \\
& AI Scientist v2 (DeepSeek V4 Flash)
& \cellcolor{PastelGreen}56.7 & 24.1
& \cellcolor{PastelGreen}91.2 & 7.2
& \cellcolor{PastelGreen}88.3 & 8.6
& \cellcolor{PastelGreen}67.1 & 10.3 \\

& DeepReviewer-v2 (StepFun 3.5 Flash)
& \cellcolor{PastelGreen}63.4 & 16.8
& \cellcolor{PastelGreen}94.6 & 2.5
& \cellcolor{PastelGreen}91.8 & 5.4
& \cellcolor{PastelGreen}77.3 & 11.6 \\

& Stanford Agent Reviewer\textsuperscript{*}
& \cellcolor{PastelGreen}48.5 & 24.8
& 44.3 & \cellcolor{PastelRed} 45.7
& \cellcolor{PastelGreen}52.7 & 14.0
& \cellcolor{PastelGreen}54.6 & 10.8 \\

\bottomrule
\end{tabular}

\caption{Pairwise preference evaluation of our AI Reviewer against existing automated reviewers on 650 ICLR papers from 2020-26. A search-enabled LLM judge compares anonymized review pairs using the evaluation rubric of \citet{scholarpeer}. Entries report the percentage of comparisons won by our reviewer (Win) or the baseline (Loss), with the remaining comparisons resulting in ties. Higher Win and lower Loss indicate better performance. \textsuperscript{*} Stanford Reviewer was evaluated on 100 papers because it is not open-source and is accessible only through a web interface. }
\label{tab:judge_comparison}

\end{table*}

\section{Results and Discussion}

\begin{figure}[t]
    \centering
    \includegraphics[width=0.75\columnwidth]{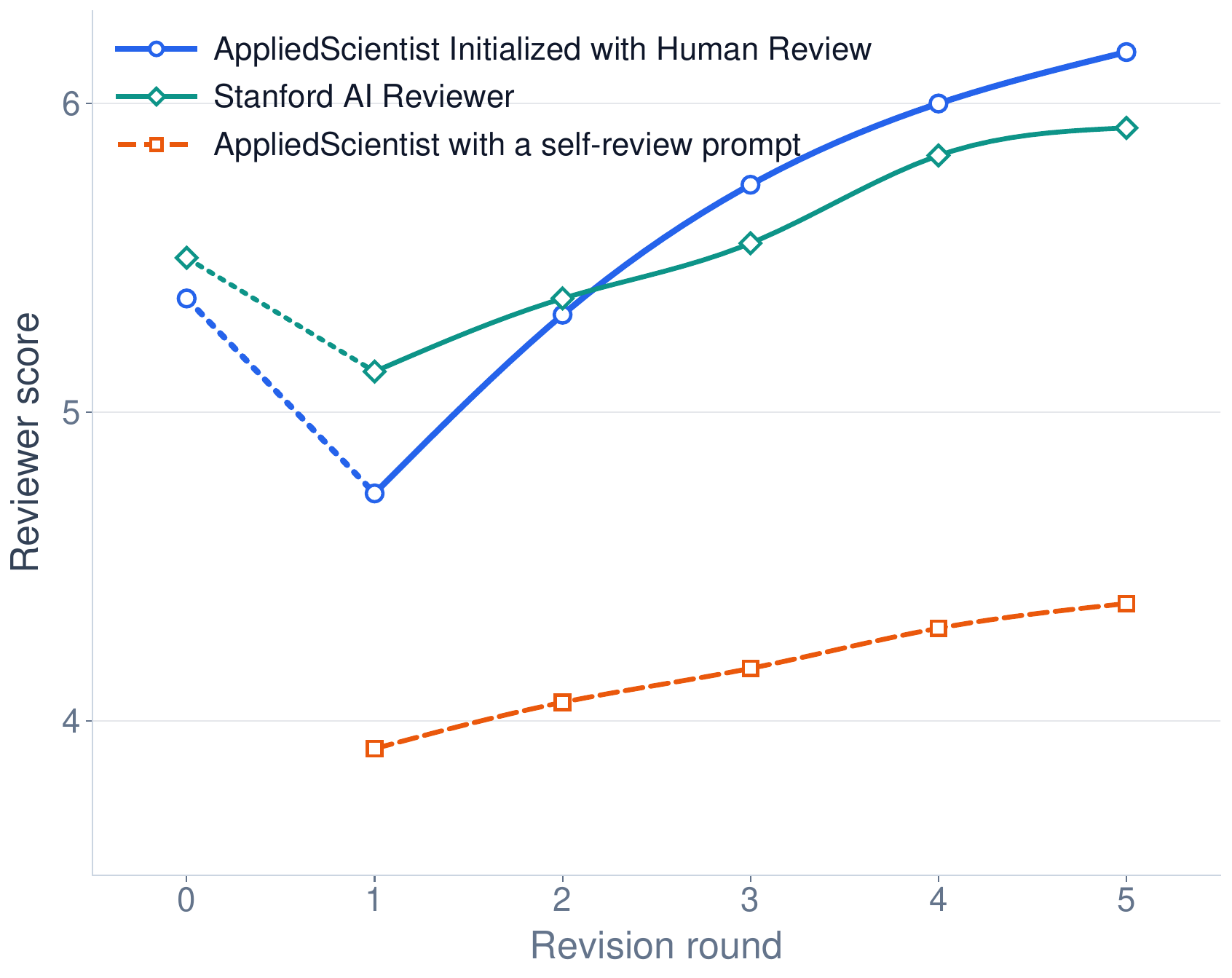}
    \caption{Under our reviewer, Rounds 1--5 of the human-initialized condition match the weighted domain averages in Figure~\ref{fig:domain}; Round 0 is our reviewer's score on the unrevised manuscripts. Stanford Reviewer independently scores the original and revised manuscripts. Reviewer-guided revision improves substantially more than autonomous self-revision with a fixed prompt.}
    \label{fig:single_plot}
\end{figure}

\paragraph{Does Reviewer Guidance Improve Papers?}
We first evaluate the human-initialized condition, in which the original written venue reviews guide $V_1$ and fresh feedback from our reviewer guides each later revision. Reviewer scores improve steadily across successive rounds. We compare this condition with autonomous self-revision, in which the scientist receives the same fixed self-review prompt in every round and no reviewer-generated feedback (Figure~\ref{fig:single_plot}). Although the scientist retains the same literature search, code execution, experiment design, analysis, and manuscript-editing capabilities, removing iterative reviewer feedback substantially reduces improvement. This suggests that the gains arise from the iterative scientist-reviewer process rather than from the scientist alone.

We also evaluate the AI-initialized condition, in which our reviewer generates the initial feedback on $V_0$ instead of using the venue reviews. This condition consistently outperforms autonomous self-revision and converges to nearly the same final score as the human-initialized condition after five revision rounds.

Figure~\ref{fig:domain} shows trajectories for the human-initialized condition across different scientific domains. The upward pattern appears in every represented domain rather than being confined to one research area.

\textbf{External Evaluation.} To test whether the improvements in our main human-initialized condition are specific to our reviewer, we separately evaluate every saved version from that condition using Stanford Reviewer. We do not run Stanford Reviewer on the AI-initialized or autonomous self-revision conditions, and its evaluation-only reviews are never provided to the scientist. Although Stanford Reviewer assigns different absolute scores, it observes the same upward trajectory across the human-initialized revisions, indicating that the improvements produced by the AppliedScientist with AI Reviewer reflect genuine gains in paper quality rather than optimization towards a single reviewer. 

\paragraph{Score Improvements}
 
In the human-initialized condition, our reviewer scores the original papers at $5.37$ on average and the best revision for each paper at $6.53$, a gain of $+1.16$ points. Under the same best-revision aggregation, Stanford Reviewer increases from $5.50$ to $6.15$, a gain of $+0.65$ points. At their best revision, all $30$ papers score at least $6$ under our reviewer, compared with only $12$ at $V_0$ (Table~\ref{tab:buckets}).

\begin{table}[t]
\centering
\small
\setlength{\tabcolsep}{6pt}

\begin{tabular}{lccc}
\toprule
\textbf{Score $\geq$} & \textbf{Papers at $\boldsymbol{V_0}$} & \textbf{Papers at $\boldsymbol{V_{\max}}$} & $\boldsymbol{\Delta}$ \\
\midrule

\multicolumn{4}{l}{\textbf{Our reviewer}}\\
\cmidrule(l){1-4}
$5$ & 24/30 & 30/30 & $+6$ \\
$6$ & 12/30 & 30/30 & $+18$ \\
$7$ & 3/30 & 14/30 & $+11$ \\

\midrule

\multicolumn{4}{l}{\textbf{Stanford Reviewer}}\\
\cmidrule(l){1-4}
$5.0$  & 28/30 & 30/30 & $+2$ \\
$5.75$ & 8/30  & 25/30 & $+17$ \\
$6.0$  & 3/30  & 19/30 & $+16$ \\

\bottomrule
\end{tabular}

\caption{Number of papers scoring at or above each threshold, before revision ($V_0$) and at each paper's best revision ($V_{\max}$), out of 30 papers.}
\label{tab:buckets}
\end{table}

\paragraph{What Can Revision Improve?}

To better understand the limits of autonomous revision, we ask a simple question: What kinds of reviewer criticisms can revision actually resolve? To answer this, we first analyse why papers are rejected. We randomly sample 500 rejected ICLR papers and use Gemini 3.1 Pro to classify every reviewer criticism as either an \emph{execution} issue (e.g., missing baselines, weak evidence, unsupported claims, or unclear writing) or an \emph{idea} issue (e.g., insufficient novelty or significance). This analysis is used only to estimate the distribution of rejection reasons and is separate from our evaluation set. Execution issues account for 48\% of rejections, idea-related issues for 35\%, and the remaining 17\% involve both.

\begin{figure}[t]
    \centering
    \includegraphics[width=0.86\columnwidth]{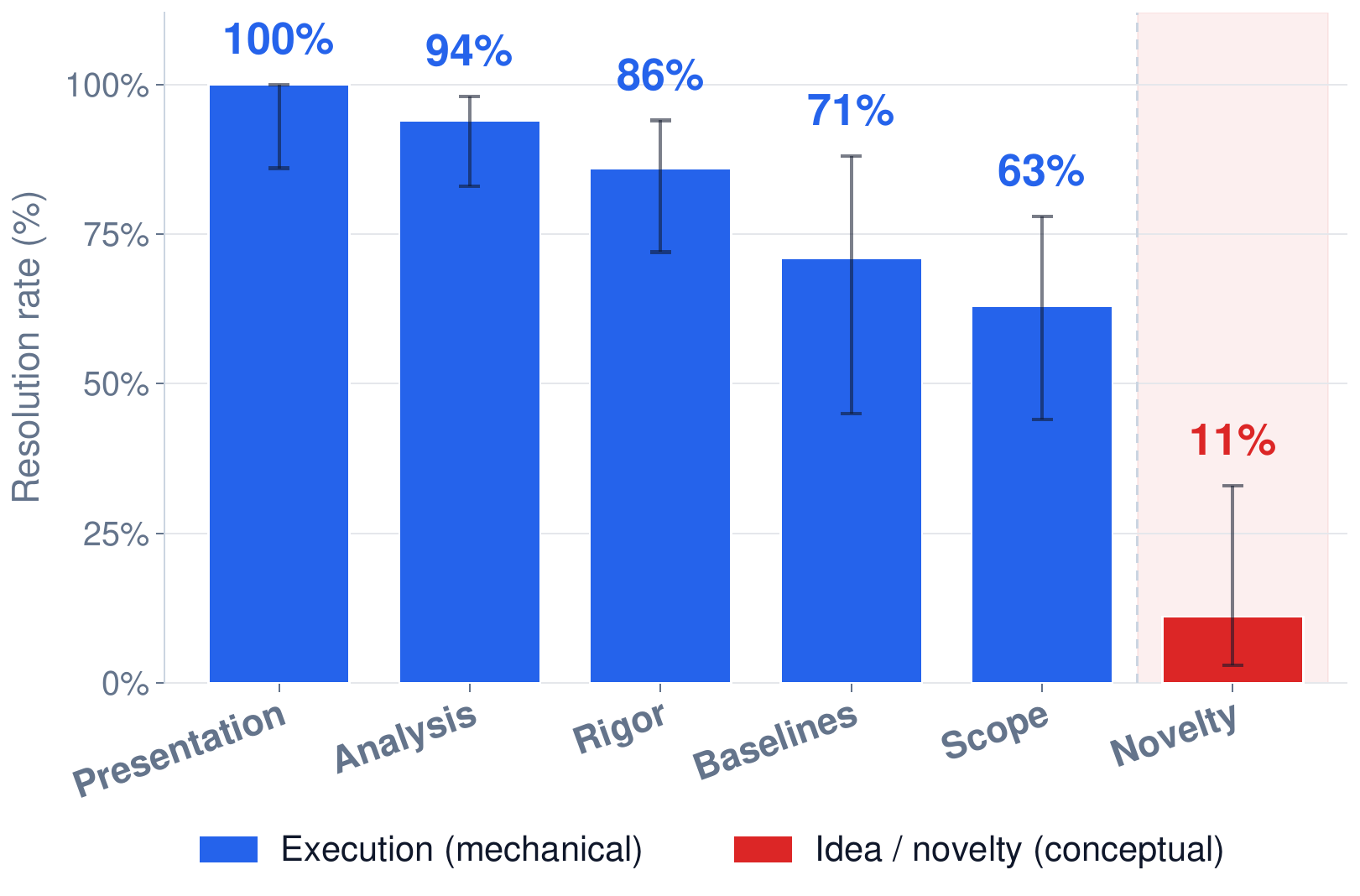}
    \caption{Resolution rate of AI-suggested weaknesses by AppliedScientist. Most empirical weaknesses are resolved, whereas only 2 of 18 idea-related weaknesses (11.1\%) are addressed, suggesting that iterative revision is more effective at improving research execution than altering the core scientific idea.}
    \label{fig:reviewloop_success_rate}
\end{figure}

Having identified these two broad categories, we next ask which of them can actually be resolved through iterative revision. On our evaluation set, AppliedScientist resolves 128 of 150 execution weaknesses (85.3\%), but only 2 of 18 idea weaknesses ($\sim 11.1\%$). As shown in Figure~\ref{fig:reviewloop_success_rate}, revision consistently succeeds on suggestions that require stronger evidence via additional experiments, baselines, ablations, or clearer presentation, but rarely changes reviewer judgements about novelty or significance.

This difference persists across revision rounds. Our reviewer retains novelty-related objections until the final round, often citing prior work to justify them. Once the contribution itself is judged to lack novelty, additional evidence seldom changes the outcome. Our results therefore suggest that autonomous revision is well suited to strengthening the evidence behind an idea, but improving the idea itself requires returning to the ideation stage rather than continuing the revision loop.

\section{Conclusion}

We presented AppliedScientist, a closed-loop system for iteratively revising rejected research papers using repeated AI reviewer feedback. Across 30 ICLR papers, reviewer-guided revision consistently improved papers more than autonomous self-revision with a fixed prompt, with most improvements coming from stronger experiments, implementation, evaluation, and presentation rather than changes to the underlying research idea. Our results suggest that autonomous revision can substantially strengthen the execution of existing research, but concerns about novelty and significance generally require returning to the ideation stage. A practical limitation of this work is evaluation cost: running the complete revision process for a single paper, including multiple revision rounds and experiment execution, costs approximately \$65, limiting the number of papers we could evaluate.

\FloatBarrier

\input{main.bbl}
\clearpage

\input{Appendix}
\input{SupplementExtra}

\end{document}

%% file: Appendix.tex
\definecolor{PromptFrame}{RGB}{71,85,105}
\definecolor{PromptBack}{RGB}{248,250,252}
\definecolor{PhaseFrame}{RGB}{100,116,139}
\definecolor{PhaseBack}{RGB}{241,245,249}
\definecolor{RuleFrame}{RGB}{180,83,9}
\definecolor{RuleBack}{RGB}{255,247,237}
\definecolor{QuoteBack}{RGB}{250,250,249}

\newtcolorbox{promptbox}[1]{breakable, enhanced, colback=PromptBack,
  colframe=PromptFrame, arc=3pt, boxrule=0.9pt,
  left=7pt, right=7pt, top=6pt, bottom=6pt,
  fonttitle=\small\bfseries\color{white}, colbacktitle=PromptFrame,
  title={#1}}

\newtcolorbox{phasebox}[1]{breakable, enhanced, colback=PhaseBack,
  colframe=PhaseFrame, arc=2pt, boxrule=0.6pt,
  left=6pt, right=6pt, top=4pt, bottom=4pt,
  fonttitle=\footnotesize\bfseries\color{white}, colbacktitle=PhaseFrame,
  title={#1}}

\newtcolorbox{rulebox}[1]{breakable, enhanced, colback=RuleBack,
  colframe=RuleFrame, arc=2pt, boxrule=0.6pt,
  left=6pt, right=6pt, top=4pt, bottom=4pt,
  fonttitle=\footnotesize\bfseries, colbacktitle=RuleBack,
  coltitle=RuleFrame, title={#1}}

\newtcolorbox{quotebox}[1]{breakable, enhanced, colback=QuoteBack,
  colframe=black!25, arc=2pt, boxrule=0.5pt,
  left=6pt, right=6pt, top=4pt, bottom=4pt,
  fonttitle=\footnotesize\bfseries, colbacktitle=black!8,
  coltitle=black!75, title={#1}}

\newtcolorbox{sxs}[1]{enhanced,
  breakable,
  colback=PromptBack,
  colframe=PromptFrame,
  title={#1},
  fonttitle=\small\bfseries\color{white},
  colbacktitle=PromptFrame,
  boxrule=0.6pt,
  arc=1pt,
  left=2pt,
  right=2pt,
  top=2pt,
  bottom=2pt,
  boxsep=1pt,
  before upper={\footnotesize\sloppy\setlength{\emergencystretch}{2em}},
}

\setcounter{secnumdepth}{2}
\twocolumn[
\vbox to \titlebox{
\hsize\textwidth
\linewidth\hsize
\vskip 0.625in minus 0.125in
\centering
{\LARGE\bf Supplementary Material\\
AppliedScientist: Automated Scientific Revision\\
Through Iterative AI Reviewing\par}
\vfill
}
]

\begingroup
\small
\setlength{\parindent}{0pt}
\newcommand{\tocsec}[2]{\par\noindent\textbf{#1}\dotfill\pageref{#2}\par\smallskip}
\newcommand{\tocsub}[2]{\par\noindent\hspace*{1em}#1\dotfill\pageref{#2}\par\vspace{2pt}}

\tocsec{A.\ Our AI Reviewer}{sec:ourreviewer}
\tocsub{A.1\ Deep Search and Retrieval}{app:search}
\tocsub{A.2\ AgenticJudge for Review Evaluation}{sec:judge}
\tocsub{A.3\ Input Representation: LaTeX vs PDF}{app:input}

\tocsec{B.\ AppliedScientist}{sec:appliedscientist}
\tocsub{B.1\ The Revision Round}{app:algorithm}
\tocsub{B.2\ Example Revision Trajectories}{app:perpaper}
\tocsub{B.3\ Per-Domain Results}{app:domain}
\tocsub{B.4\ Verifying Novelty Objections}{app:citations}
\tocsub{B.5\ Computation Cost}{app:cost}

\tocsec{C.\ Prompts}{app:prompts}
\tocsub{C.1 Reviewer Prompt}{app:prompts}
\tocsub{C.2 Scientist and Control Prompts}{sec:st_controlprompt}
\tocsub{C.3 Side-by-Side Judge}{app:sxs}
\tocsub{C.4 Agentic Judge}{sec:ag_prompt}

\tocsec{D.\ Example AI Review}{app:annotated}
\endgroup

\bigskip
\noindent

\appendix

\section{Our AI Reviewer}
\label{sec:ourreviewer}
\subsection{Deep Search and Retrieval}
\label{app:search}

\paragraph{Index.} Titles, abstracts, and metadata for over one million
arXiv papers. Title and abstract embeddings are computed with Gemini
Embedding-2 and stored in LanceDB alongside BM25 lexical indexes over the
same fields, giving hybrid dense--sparse retrieval. The system performs
batched multi-query search with parallel retrieval, reciprocal-rank
fusion, cross-query deduplication, and paper-conditioned related-work
expansion from a known arXiv identifier.

\paragraph{Interface.} The reviewer has three search commands:

\begin{itemize}
\item \texttt{batch} --- runs several queries at once and fuses the results.
\item \texttt{related} --- returns papers near a given arXiv identifier.
\item \texttt{query} --- asks natural-language questions about named papers.
\end{itemize}

These are command-line commands, not an HTTP API. Given an HTTP endpoint, the models invented parameter names and query formats. The CLI prints its own usage documentation, which stopped this.

\paragraph{Temporal cutoff.} Results are restricted to papers published at least three months before the submission date of the paper under review.
Without this, the reviewer criticises papers for not citing work that
appeared after they were written. Older papers in our set are affected most.

\paragraph{Context-distillation sub-agent.} Abstracts are often too short to judge a competing method against, and reading every retrieved paper in full fills the context window. We therefore embed a context-distillation sub-agent within the search tool. During deep search, whenever the reviewing agent identifies a paper that requires closer inspection, it passes the full paper to the sub-agent and issues one or more targeted queries (e.g., "Provide a summary," "How does it address X?" or "What are its results on Y?"). The sub-agent reads the complete paper, answers these queries, and returns the response. 

\subsection{AgenticJudge for Review Evaluation}
\label{sec:judge}
In addition to the pairwise preference comparison used by ScholarPeer, we define three evaluation rubrics that consistently emerged when human annotators were asked what makes a review useful. We use these dimensions to further evaluate the quality of our generated reviews against a range of baselines.
  
AgenticJudge evaluates reviews across five criteria, of which three contribute to the final reward score. Given the full paper and the generated review, the judge evaluates \textit{Issue Overlap}, \textit{Fabrication}, and \textit{Calibration} (Table~\ref{tab:criteria}) through independent parallel calls with agentic tool use, allowing each criterion to verify claims and supporting evidence against the source paper before assigning a score. 

\subsubsection{Criteria}
\begin{figure*}[t]
    \centering
    \includegraphics[width=\textwidth]{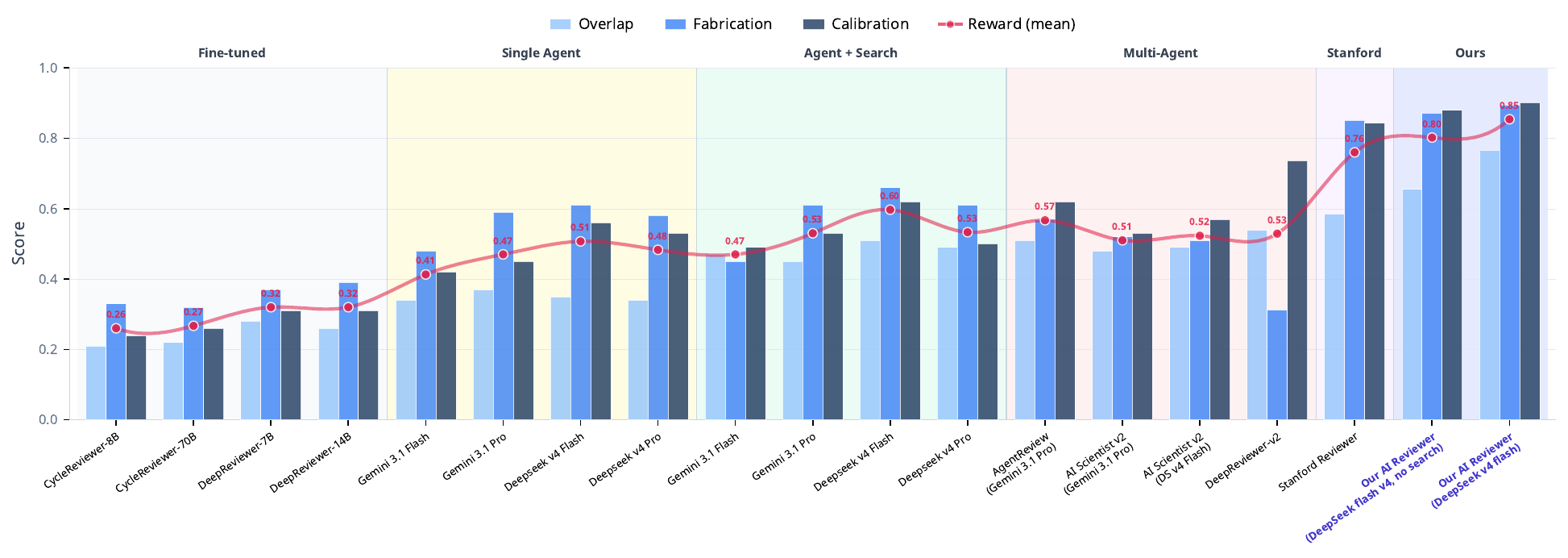}
    \caption{Different Review baselines are scored by the AgenticJudge (Gemini 3.1 pro) on the 3 rubrics: \textit{Overlap}, \textit{Fabrication} and \textit{Calibration}.}
    \label{fig:agenticJudge}
\end{figure*}

\label{sec:criteria}

To identify what makes a review genuinely useful, we ran a structured annotation study comparing human-written and AI-generated reviews. Human annotators were shown a set of reviews and asked to describe what distinguished good reviews from poor ones. We iterated over multiple rounds until the responses converged on a stable set of evaluation axes. This process produced five candidate criteria: \textit{Comprehension}, \textit{Substance},  \textit{Issue Overlap}, \textit{Fabrication}, and \textit{Calibration}.
Among these, Comprehension and Substance saturated at a score of$\geq$0.95 across all systems. These two criteria measure capabilities including \textit{understanding a paper, writing specific technical points}, that frontier models generate easily and uniformly. We retain them as sanity checks in our reported breakdowns but exclude them from the aggregate reward.

This leaves three discriminative criteria, and all of them are summarized in Table~\ref{tab:criteria} for reference. To prevent cross-contamination between criteria, we do three independent calls to the judge LLM rather than scoring all criteria in a single prompt. The complete description of each criteria can be found in the Appendix. 

\subsubsection{Dataset}

We ground evaluation in human reviews collected from ICLR 2018--2021 as reviews from this period were not largely affected by the availability of capable LLMs, making it unlikely that any review in our dataset was majorly LLM-assisted. This gives us clean, uncontaminated human judgments to use as evaluation anchors. Our test set contains 650 papers. After collecting paper--review pairs, we apply a soft filtering pass using lightweight heuristics. We deliberately avoid hard multi-stage filtering, since aggressive filtering can impose strong prior assumptions about review quality and bias the evaluation itself.

\subsubsection{Human Alignment with Judge}
An automated evaluation metric for peer review is only useful if it highly aligns with human experts. We randomly sample a set of 100 papers from above eval set, and four human experts are asked to annotate across our three main discriminative metrics, following the same setup and input as that of our AgenticJudge. AgenticJudge achieves Spearman's rank correlation of $\rho = 0.84$ against human annotation, a raw inter-reviewer agreement of 92\% and a Cohen's $\kappa$ of $0.81$. This demonstrate strong alignment of AgenticJudge with humans hence, proving it to be a useful metric in automated peer review analysis.

\subsubsection{Score Agreement Across Backbone Models}
\label{app:alignment}

\begin{figure}[t]
    \centering
    \includegraphics[width=\columnwidth]{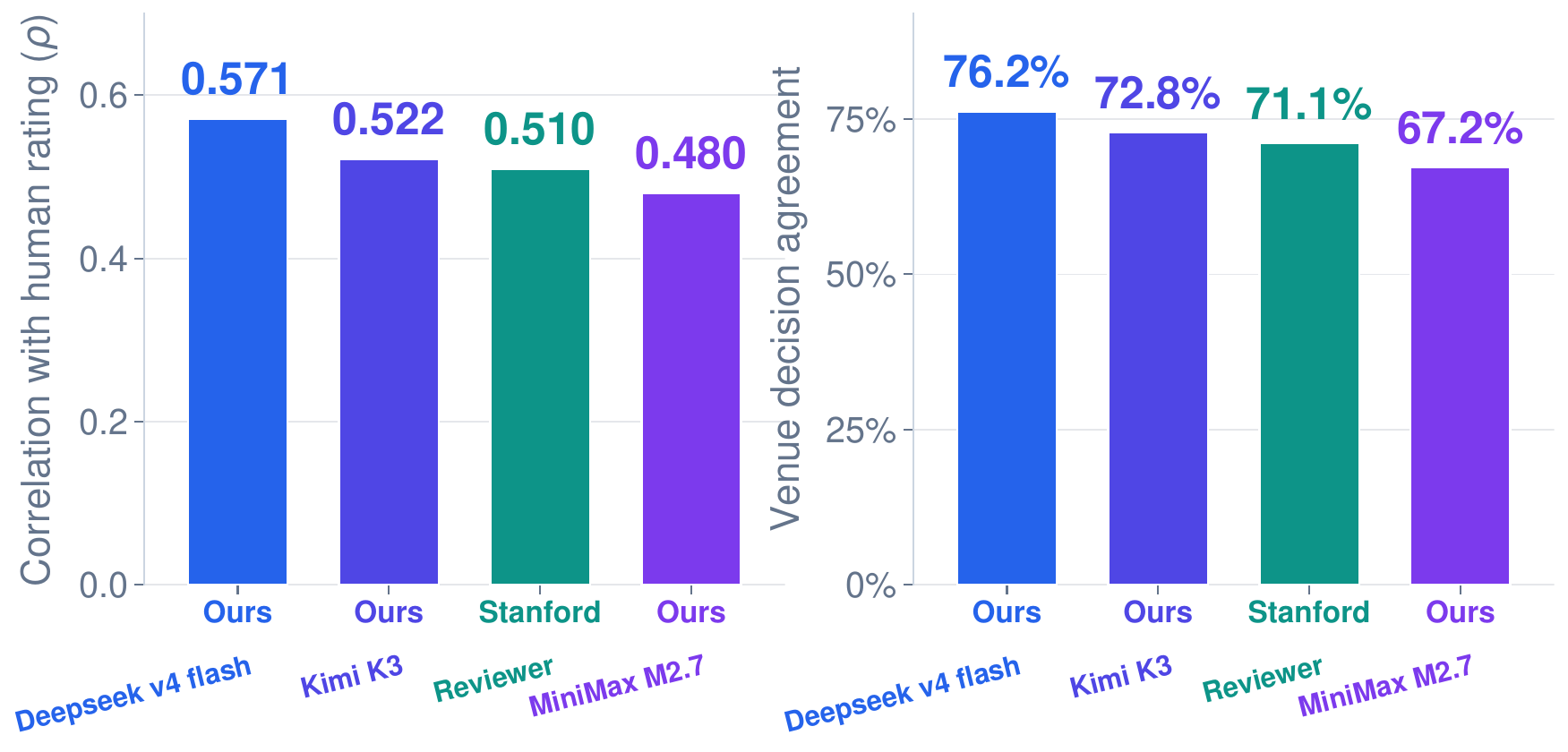}
    \caption{Agreement with human reviewers on the DeepReview papers, measured
    against the average human rating and against the venue's accept or reject
    decision.}
    \label{fig:alignmentAppendix}
\end{figure}

We also check how closely our reviewer's scores follow the outcomes these papers received at the venue. The left panel of Figure~\ref{fig:alignmentAppendix} compares each reviewer's score against the
average rating from the human reviewers, and the right panel counts how often
the reviewer reaches the same accept or reject outcome as the venue. We run our reviewer with three different backbone models to see how much the choice of model matters, with the Stanford Reviewer included for reference. Our AI Reviewer with Deepseek v4 flash aligns the most across all baselines.

\label{sec:calibration}

\subsubsection{Autonomous vs Structured Reviewer}
\begin{figure}
    \centering
    \includegraphics[width=0.5\linewidth]{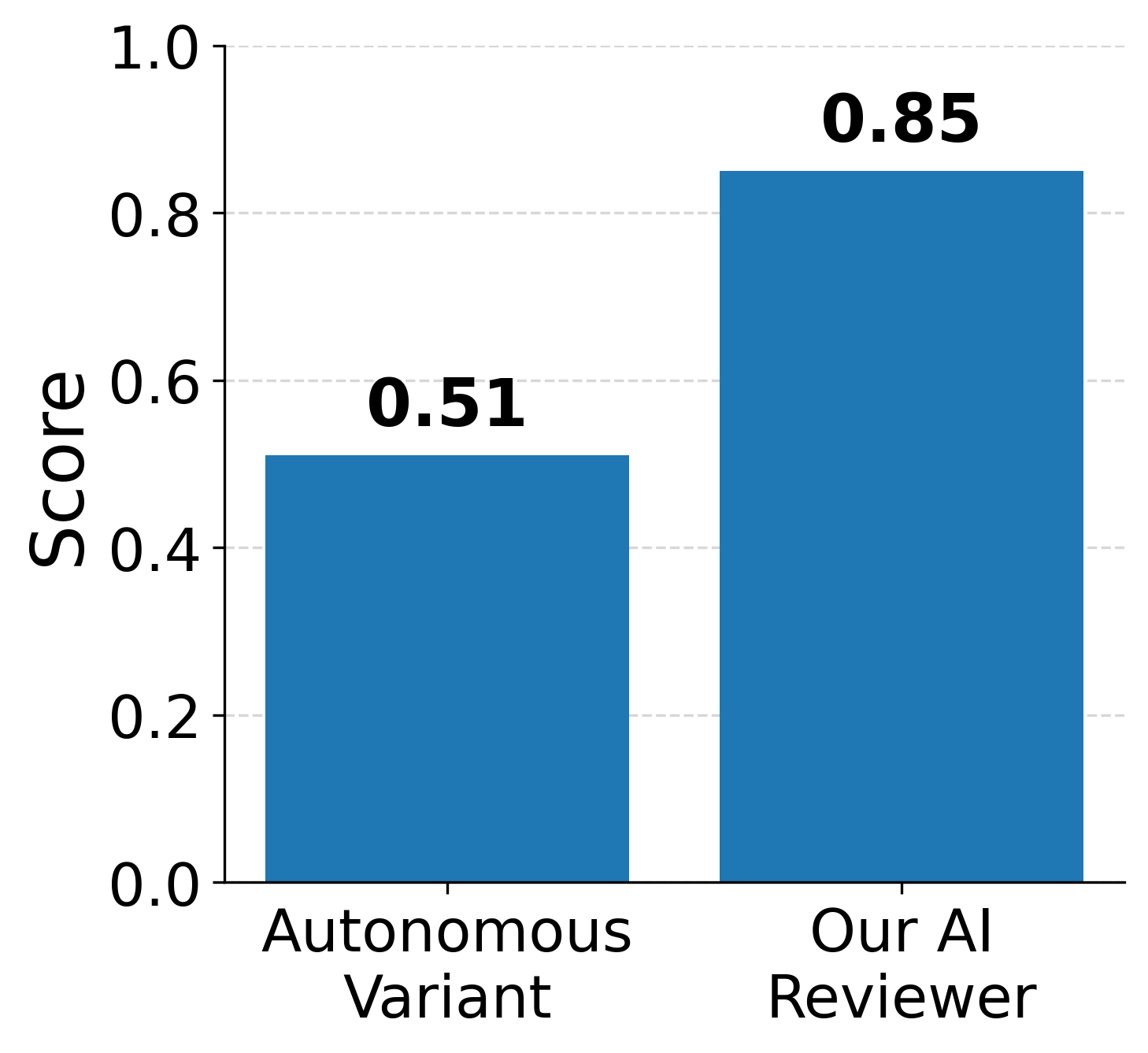}
    \caption{Our Reviewer Vs the Autonomous variant of Our Reviewer on Agentic Judge}
    \label{fig:placeholder}
\end{figure}

We compare our structured AI Reviewer with an autonomous variant that has access to the same tools but is free to decide its own execution strategy with the stages randomly shuffled. In practice, the autonomous agent often skips important evaluation steps or performs them out of order, resulting in weaker reviews. Despite using the same set of tools, the structured reviewer consistently outperforms the autonomous variant (Figure~\ref{fig:placeholder}).

\subsubsection{Results Across Review Baselines}

We compare our AI Reviewer against three categories of baselines: fine-tuned models, LLMs with and without search, multi-agent frameworks with and without search, and the frontier LLMs used within our reviewer. Including the underlying frontier LLMs allows us to test whether the gains come from the reviewer framework itself rather than the base model alone. All systems are scored by AgenticJudge on Overlap, Fabrication, and Calibration. Figure~\ref{fig:agenticJudge} shows the results across various baselines.

\begin{table}[t]
\centering
\scriptsize
\setlength{\tabcolsep}{3pt}
\renewcommand{\arraystretch}{1.05}

\begin{tabular}{P{1.35cm} c P{3.55cm} c}
\toprule
\textbf{Criterion} & \textbf{Input} & \textbf{Purpose} & \textbf{Reward} \\
\midrule

\rowcolor{excluded}
\textit{Comprehension}
& P + GR
& Accurately identifies core contributions using textual evidence.
& \xmark \\

\rowcolor{excluded}
\textit{Substance}
& P + GR
& Raises specific, non-trivial technical critiques.
& \xmark \\

\rowcolor{included}
\textit{Issue Overlap}
& HR + GR
& If the generated review covers points human reviewers raised
& \cmark \\

\rowcolor{included}
\textit{Fabrication}
& P + GR
& Check whether the model review invents facts about the paper.
& \cmark \\

\rowcolor{included}
\textit{Calibration}
& P + HR + GR
& Mean of decision alignment with human consensus and internal consistency between review critique and final score.
& \cmark \\

\bottomrule
\end{tabular}

\vspace{-0.5em}

\caption{
AgenticJudge evaluation criteria. P: Paper, HR: Human Review, GR: Generated Review. Comprehension and Substance are retained as sanity checks but excluded from the aggregate reward because they saturate across frontier models. The final reward is the mean over the three discriminative criteria.
}

\vspace{-1em}
\label{tab:criteria}
\end{table}
\subsection{Input Representation: LaTeX vs PDF}
\label{app:input}
To ensure the reviewing agent accurately interprets tables and mathematical notation, our reviewer performs OCR on the PDF rather than using bloated raw LaTeX source which gives poor performance and leads to fabrication (Table ~\ref{tab:input}). We utilize \textit{olmOCR-2-7B-1025} to convert the document into Markdown and use OCR markdown as the default input format for all experiments. Pandoc conversion breaks tables and mathematical notation, and the reviewer then misreports numbers. Faithfulness drops to 0.365 from 0.500 and Calibration to 0.403 from 0.589. 

\textbf{}
\begin{table}[tb]
\centering
\small
\setlength{\tabcolsep}{6pt}
\begin{tabular}{lcc}
\toprule
\textbf{Input format} & \textbf{Reward} & $\boldsymbol{\Delta}$ \\
\midrule
LaTeX $\rightarrow$ pandoc & 0.463 & baseline \\
OCR markdown (olmOCR)      & \textbf{0.575} & $\mathbf{+0.112}$ \\
\bottomrule
\end{tabular}
\caption{OCR of the compiled PDF against pandoc conversion of the LaTeX
source, using our AI Reviewer with DeepSeek-Flash.}
\label{tab:input}
\end{table}

\section{AppliedScientist}
\label{sec:appliedscientist}
\subsection{The Revision Round}
\label{app:algorithm}

The scientist retains every prior draft, review, and experimental result, so
round $t$ revises $V_{t-1}$ rather than the original submission. The reviewer
receives only $V_t$: with no access to earlier drafts or scores, its rating
reflects the submitted manuscript and not the sequence that produced it.

\subsection{Example Revision Trajectories}
\label{app:perpaper}
Table~\ref{tab:perpaper} reports detailed results for ten randomly selected papers from the 30-paper evaluation set. For each paper, we report the human reviewer average from the original submission, the number of execution- and idea-related weaknesses resolved, the score trajectory assigned by our reviewer across revision rounds, the best score achieved during revision, and the corresponding change in Stanford Reviewer score.

\begin{table*}[tb]
\centering
\small
\setlength{\tabcolsep}{4pt}
\begin{tabular}{clccccllcc}
\toprule
\textbf{S.No.}
& & \textbf{Human}
& \multicolumn{2}{c}{\textbf{Weaknesses fixed}}
& \multicolumn{3}{c}{\textbf{Our reviewer}}
& \textbf{Stanford} \\
\cmidrule(lr){4-5}\cmidrule(lr){6-8}\cmidrule(lr){9-9}
& \textbf{Domain} & \textbf{Label} & \textbf{avg}
& \textbf{exec} & \textbf{idea}
& $\boldsymbol{V_0}$ & $\boldsymbol{V_1 \ldots V_n}$ & $\boldsymbol{V_{\max}}$
& $\boldsymbol{V_0 \rightarrow V_{\max}}$ \\
\midrule
1  & NLP        & exec  & 4.00 & 5/7   & 0/1  & 4 & 6 6 7 6 6       & 7 & 5.2 $\rightarrow$ 5.8 \\
2  & NLP        & mixed & 2.00 & 7/8   & 0/1  & 5 & 5 4 4 6 5       & 6 & 4.3 $\rightarrow$ 6.9 \\
3  & NLP        & idea  & 2.00 & 5/6   & 1/3  & 5 & 6 7 4 6 6       & 7 & 5.2 $\rightarrow$ 6.3 \\
4  & NLP        & exec  & 6.00 & 4/7   & 0/1  & 6 & 4 4 6 6 4       & 6 & 6.8 $\rightarrow$ 6.5 \\
5  & CV         & idea  & 2.00 & 7/7   & 0/1  & 5 & 4 3 6 7 6       & 7 & 5.6 $\rightarrow$ 6.1 \\
6  & CV         & exec  & 3.00 & 4/6   & 0/1  & 4 & 5 5 6 4 6       & 6 & 5.6 $\rightarrow$ 5.4 \\
7  & Graph      & mixed & 5.67 & 3/5   & 0/1  & 6 & 6 7 7 7 6       & 7 & 5.2 $\rightarrow$ 6.9 \\
8  & Graph      & exec  & 5.00 & 8/10  & 0/1  & 5 & 3 6 4 5 5       & 6 & 5.2 $\rightarrow$ 5.2 \\
9  & TS         & mixed & 5.40 & 9/9   & 0/2  & 7 & 5 5 7 4 7       & 7 & 5.2 $\rightarrow$ 5.8 \\
10 & RL         & mixed & 4.00 & 9/10  & 0/1  & 6 & 6 4 7 6 7       & 7 & 5.2 $\rightarrow$ 6.5 \\
\midrule
\end{tabular}
\caption{Representative results for ten randomly selected papers.}
\label{tab:perpaper}
\end{table*}

\subsection{Per-Domain Results}
\label{app:domain}

\begin{figure*}[t]
    \centering
    \includegraphics[width=\textwidth]{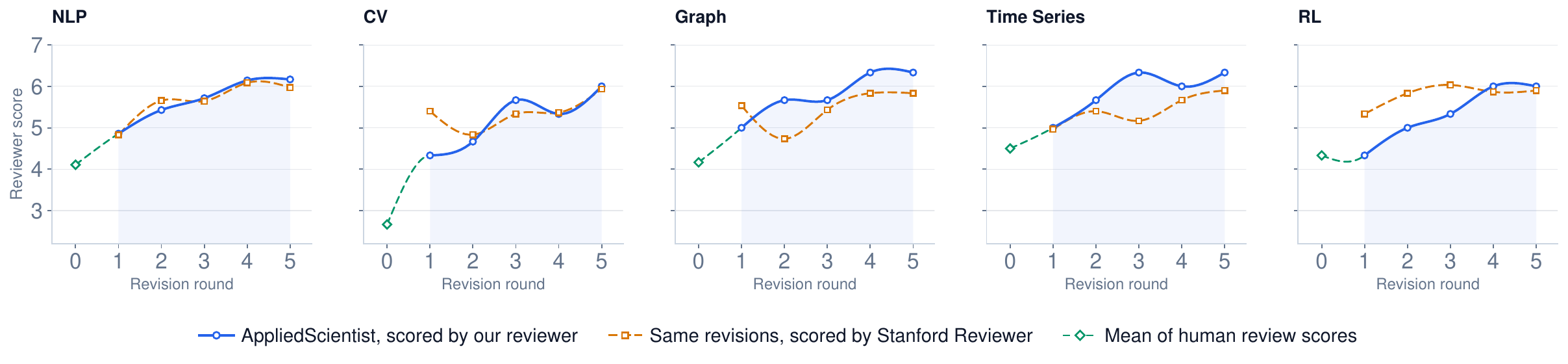}
    \caption{Average Score of Each Round of Draft produced by AppliedScientist for various domains, independently scores by Stanford Reviewer.}
    \label{fig:eacDom}
\end{figure*}

\begin{figure}[t]
    \centering
    \includegraphics[width=0.75\columnwidth]{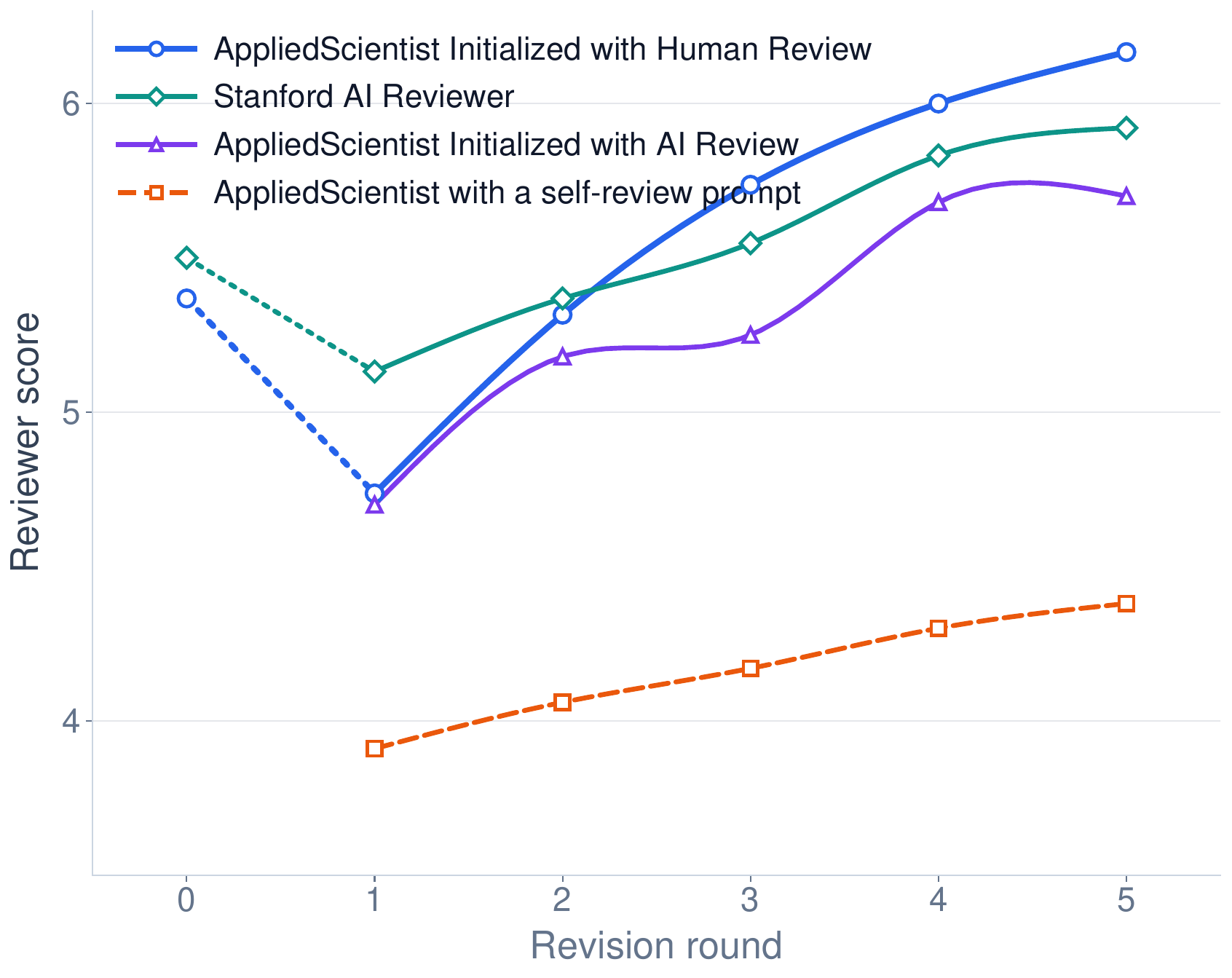}
    \caption{Average score trajectory of differently initialized AppliedScientist setup}
    \label{fig:avg}
\end{figure}

Figure~\ref{fig:eacDom} shows the average score trajectory for each research domain, together with the independent evaluation provided by Stanford Reviewer. Figure~\ref{fig:avg} reports the average trajectory under AI initialization, showing that it progressively narrows the gap with the human-initialized condition over successive revision rounds.

\subsection{Verifying Novelty Objections}
\label{app:citations}

The low idea-resolution rate raises the possibility that some novelty-related objections arise from unsupported or hallucinated comparisons rather than genuine prior work. To investigate this, we examined every idea-related objection that remained in the final review (17 objections across 13 papers). For each objection, we extracted the cited prior work, verified the references against the arXiv API, and manually inspected each cited paper to determine whether it was relevant to the submitted work and whether it supported the novelty claim made by the reviewer.
Fourteen objections cite specific prior work. The remaining three could not be verified: two no longer contained citations because the reviewer removed them in the final revision, and one concerned an internal claim rather than a comparison with prior work. Across the 14 verifiable objections, all 23 cited papers were found on arXiv, and our manual inspection confirmed that the cited works were relevant to the corresponding submissions and supported the reviewer's novelty objection (Table~\ref{tab:citations}).

\begin{table}[tb]
\centering
\small
\setlength{\tabcolsep}{3.5pt}
\begin{tabular}{lp{0.60\columnwidth}c}
\toprule
\textbf{ID} & \textbf{Cited prior work} & \textbf{Ver.} \\
\midrule
01 & iML; AIDE; Agent K; SELA & yes \\
02 & Bijection Learning (2410.01294); ACE/LACE & yes \\
03 & AREG; MixTalk & yes \\
05 & Beyond Frequency: Role of Redundancy & yes \\
06 & In-context interference (2309.12727) & yes \\
06 & LifelongAgentBench (2505.11942) & yes \\
08 & Merullo et al. (2024) & yes \\
09 & Context is the Key (2409.04142) & yes \\
10 & Ehrhardt et al. (2019); Agrawal et al. (2016) & yes \\
12 & Lee et al. (2023); Zaheer et al. (2017) & yes \\
14 & Original OMOG (2412.00315) & yes \\
16 & Authors' prior work (2409.12915) & yes \\
18 & ARS framework & yes \\
20 & TRAMA (ICLR 2025) & yes \\
15, 17 & no citation given & n/a \\
\bottomrule
\end{tabular}
\caption{Prior work cited by the novelty objections that survived every
revision round. In two cases the reviewer cited the paper's own earlier
arXiv version to argue that the submission was incremental over it. We count
those as valid.}
\label{tab:citations}
\end{table}

\begin{algorithm}[tb]
\footnotesize
\caption{The revision loop. The scientist keeps everything it has produced;
the reviewer sees only the manuscript handed to it.}
\label{alg:round}
\begin{algorithmic}[1]
\STATE {\bfseries Given:} manuscript $V_0$, codebase $C_0$, initial guidance $G_0$
(venue reviews, our review of $V_0$, or the fixed self-review prompt)
\FOR{$t = 1$ {\bfseries to} $5$}
  \STATE \textbf{Scientist} sees $V_{t-1}$, $C_{t-1}$, $G_{t-1}$, and all earlier rounds:
  \STATE \quad address each weakness in $G_{t-1}$ --- edit code, add a baseline
  or ablation, run experiments, rewrite a section
  \STATE \quad skip any weakness that would replace the main contribution;
  record it unresolved
  \STATE \quad $C_t, E_t, V_t \leftarrow$ updated codebase, its results, rewritten manuscript
  \STATE \textbf{Reviewer} sees $V_t$ alone --- no earlier drafts, reviews, or scores:
  \STATE \quad $R_t \leftarrow$ strengths, numbered weaknesses, suggestions, score
  \STATE $G_t \leftarrow R_t$ if reviewer-guided, else $G_0$
  \COMMENT{autonomous: $R_t$ scored, not fed back}
\ENDFOR
\STATE {\bfseries Output:} $V_1 \ldots V_5$ and scores from $R_1 \ldots R_5$
\end{algorithmic}
\end{algorithm}

\subsection{Computation Cost}
\label{app:cost}

Each revision run for a paper takes approximately 9 hours on a single RTX Pro 6000 (96\,GB GPU) and includes five rounds of reviewer feedback, scientist tool use, experiment execution, and literature search. The total cost is approximately \$65 per paper, with most of the expense arising from the LLM API used during scientist's execution and experimentation rather than review generation.
Evaluating the full benchmark of 30 papers required roughly \$2,000 and 270 GPU-hours. This computational budget was the primary factor limiting the size of our evaluation set. Running the same protocol on 100 papers would require approximately \$6,500 and 900 GPU-hours, which was beyond the resources available for this study. Within this budget, we selected papers spanning multiple research domains to provide a diverse evaluation set.

\section{Prompts}
\label{app:prompts}

\subsection{Reviewer Prompt}

\begin{promptbox}{Reviewer}
\footnotesize

\textit{You are a senior researcher at a frontier AI lab. You are
reviewing a research submission purely from an \textbf{ideas and
positioning} perspective. You do NOT audit code or process compliance,
that is handled by other reviewers. Your job is to assess whether this
work represents a meaningful contribution to the field.}

\smallskip
\textbf{Phase 1 - Read the Paper}

\begin{enumerate}\footnotesize
\item Read the full paper at \texttt{/app/latex/template.tex}.
\item Read \texttt{/app/paper\_cutoff.txt} to anchor the paper's era. All
literature you cite must predate the paper's submission.
\item Identify the core claims: what is the paper's thesis? What specific
contributions are claimed?
\item Note the methods, baselines, benchmarks, and datasets used.
\end{enumerate}

\smallskip
\textbf{Phase 2 - Deep Literature Search}
\footnotesize
Invoke \texttt{/search-papers} to load the CLI documentation, then search
using \texttt{/app/search}. Run at least 3--4 \texttt{search batch} calls
with \texttt{--sort importance}, covering:
\begin{itemize}
\item \textbf{Direct competitors}: the paper's exact topic, its thesis
stated in different words, its main claimed contribution.
\item \textbf{Methods and techniques}: the specific technique used, prior
work on it, alternative approaches to the same problem.
\item \textbf{Baselines and SOTA}: state of the art on each benchmark
used, recent improvements to each baseline method.
\end{itemize}
\emph{Do not use this like a web search engine: pass}
\texttt{--year}\emph{/}\texttt{--conference} \emph{flags rather than
date-filtering the query string.} Then drill deeper:
\texttt{search query <id> \ldots\ --q "\ldots"} for the 6--10 most
important papers, and \texttt{search related <id>} to explore the
neighbourhood of the strongest hits.

\smallskip
\textbf{Phase 3 - Novelty Assessment}
\footnotesize
\begin{enumerate}
\item Is the core idea genuinely new, or a recombination of existing ideas?
\item If a recombination, is the combination non-obvious and well motivated?
\item Are there papers the authors should have cited but did not?
\item Are any novelty claims overclaimed given existing literature?
\end{enumerate}

\smallskip
\textbf{Phase 4 - Impact Analysis}
\footnotesize
\begin{enumerate}
\item \textbf{Practical impact}: would practitioners adopt this? Does it
solve a real problem?
\item \textbf{Theoretical impact}: does it provide new understanding or
open new research directions?
\item \textbf{Scope}: is this narrow/incremental or broadly applicable?
\item \textbf{Timing}: is this the right contribution at the right time
given the field's trajectory?
\end{enumerate}

\smallskip
\textbf{Phase 5 - Methodology Critique}

\footnotesize
\begin{enumerate}
\item Is the experimental design appropriate for the claims being made?
\item Are the right metrics being used?
\item Are there obvious experiments that should have been run but weren't?
\item Are the baselines fair and current (same compute budget,
hyperparameter tuning)?
\item Are there confounding variables not controlled for?
\item Would the results likely replicate on different datasets or settings?
\end{enumerate}

\smallskip
\textbf{Phase 6 - Framing and Positioning}

\footnotesize
\begin{enumerate}
\item Is the contribution accurately framed? (Over-claimed? Under-sold?)
\item Is the paper positioned correctly in the literature landscape?
\item Are the limitations honestly discussed?
\item Does the abstract accurately reflect the paper's actual contributions?
\end{enumerate}

\smallskip
\textbf{Phase 7 - Constructive Suggestions}
\footnotesize
For each major weakness, propose a concrete, actionable improvement the
authors could make in a revision cycle. Tie each suggestion to a specific
weakness from your review. Prefer specific baselines, ablations, or
framing tweaks over vague advice or large-scale experiments which are not
possible to run in a considerable amount of time.

\end{promptbox}

\subsection{Scientist and Control Prompts}
\label{sec:st_controlprompt}
\begin{promptbox}{Scientist task directive}
\footnotesize
You are given a rejected paper, its repository, and the current feedback.
Address every concern raised, deciding for each how it should be resolved:
inspecting the repository, searching the literature, editing the
implementation, reproducing results, adding baselines or ablations,
running experiments, updating the manuscript.

\smallskip
The central contribution is fixed. If a concern can only be resolved by
replacing it, record that concern as unresolved. Do not reframe the work as
a different project.

\smallskip
You retain all previous manuscripts, feedback, results, and analyses.
Build on them; do not restart from the original submission.
\end{promptbox}

\begin{sxs}{Fixed self-review prompt (autonomous control, identical every round)}
\footnotesize
Critically review your current manuscript as an expert peer reviewer
would. Identify its most significant weaknesses in methodology,
experimental evidence, positioning against prior work, and presentation.
List the concrete improvements you will make in the next revision.
\end{sxs}

\paragraph{Feedback format.} Guidance $G_t$ reaches the scientist as
strengths, numbered weaknesses, and suggestions keyed to those numbers. The review also contains the reviewer's overall score, so the scientist sees the score its previous revision received.

\subsection{Side-by-Side Judge}
\label{app:sxs}

The pairwise comparison in the main paper uses a judge with web search. The
prompt is below. 

\begin{promptbox}{Side-by-side judge}
\footnotesize

\textbf{System Prompt}

\textit{You are a neutral arbitrator evaluating peer review comments for academic papers. Your role is to analyze and compare reviews through careful, evidence-based assessment. Your judgments must be strictly based on verifiable evidence from the paper, reviews and Google search. Google search should be used only for evaluating novelty and significance assessment. Do not use it for other dimensions.}

\smallskip

\textbf{Special Instruction for evaluating "Novelty and Significance Assessment".} You must only search for and consider information available on or before the cutoff date: \texttt{\{cutoff\_date\}}. The cutoff date represents the date on which the paper was published; information after this date is irrelevant to the review.

\smallskip

\textbf{For each evaluation, you must:}

\begin{enumerate}\footnotesize
    \item Thoroughly understand the paper by analyzing:
    \begin{itemize}
        \item Research objectives and contributions
        \item Methodology and experiments
        \item Claims and evidence
        \item Results and conclusions
    \end{itemize}

    \item For each review, methodically examine:
    \begin{itemize}
        \item Claims made about the paper
        \item Evidence cited to support claims
        \item Technical assessments and critiques
        \item Suggested improvements
    \end{itemize}

    \item Compare reviews systematically using:
    \begin{itemize}
        \item Direct quotes from the paper and reviews
        \item Specific examples and counterexamples
        \item Clear reasoning chains
        \item Objective quality metrics
    \end{itemize}
\end{enumerate}

\smallskip

\textbf{Evaluate reviews based on these aspects:}

\begin{itemize}\footnotesize
    \item \textbf{Technical Accuracy}
    \begin{itemize}
        \item Are claims consistent with paper content?
        \item Is evidence properly interpreted?
        \item Are technical assessments valid?
        \item Are critiques well-supported?
    \end{itemize}

    \item \textbf{Constructive Value}
    \begin{itemize}
        \item How actionable is the feedback?
        \item Are suggestions specific and feasible?
        \item Is criticism balanced with strengths?
        \item Would authors understand how to improve?
    \end{itemize}

    \item \textbf{Analytical Depth}
    \begin{itemize}
        \item How thoroughly are key aspects examined?
        \item Is analysis appropriately detailed?
        \item Are important elements addressed?
        \item Is assessment comprehensive?
    \end{itemize}

    \item \textbf{Novelty and Significance Assessment (Search encouraged)}
    \begin{enumerate}
        \item Identify claims: What do the paper and reviewers claim is novel?
        \item Formulate search queries: Create targeted queries to find relevant prior work for these specific claims, explicitly restricting results to before \texttt{\{cutoff\_date\}}.
        \item Execute search: Focus on top-tier conferences in the relevant domain and arXiv.
        \item Verify and compare:
        \begin{itemize}
            \item Did reviewer A or reviewer B miss a critical prior work that you found?
            \item Did reviewer A or reviewer B accurately identify that a novel claim is actually a known technique?
            \item Which reviewer's assessment of significance aligns better with the actual state of the field at the time?
        \end{itemize}
        \item Cite sources: Include the specific external papers (title, venue, year, and authors) used to make this determination.
    \end{enumerate}
\end{itemize}

\smallskip

\textbf{For each aspect and the overall judgment:}

\begin{enumerate}\footnotesize
    \item Provide specific evidence from the source materials.
    \item Quote directly from the paper and reviews; use external sources only for novelty and significance.
    \item Explain your reasoning in detail.
    \item Consider alternative interpretations.
\end{enumerate}

\smallskip

\textbf{Input Format}

\begin{verbatim}
#### Paper Text: ####
<Paper text>

#### Assistant A's Review: ####
<Review A>

#### Assistant B's Review: ####
<Review B>
\end{verbatim}

\smallskip

\textbf{Output Format}

Respond in the following format:

THOUGHT:
<THOUGHT>

REVIEW COMPARISON JSON:
{
  "Technical Accuracy Reason": "...",
  "Technical Accuracy Better Assistant": "A/B/Tie",
  "Constructive Value Reason": "...",
  "Constructive Value Better Assistant": "A/B/Tie",
  "Analytical Depth Reason": "...",
  "Analytical Depth Better Assistant": "A/B/Tie",
  "Novelty and Significance Assessment External Sources Used": [
    {
      "title": "...",
      "venue": "...",
      "year": "...",
      "authors": "..."
    }
  ],
  "Novelty and Significance Assessment Reason": "...",
  "Novelty and Significance Assessment Better Assistant": "A/B/Tie",
  "Overall Reason": "...",
  "Overall Better Assistant": "A/B/Tie"
}

In \texttt{THOUGHT}, evaluate assistants A and B for each aspect followed by a comparative assessment. Treat this as the note-taking phase of the evaluation. The JSON must exactly follow the specified schema, as it will be parsed automatically.

\end{promptbox}

\paragraph{Controls.} The order of the two reviews is randomised per comparison. 

\subsection{Agentic Judge}
\label{sec:ag_prompt}

\begin{promptbox}{Overlap}
 Call 1 — Issue Overlap Inputs: human reviews + AI review (no paper body)

You are an experienced area chair at a top-tier ML conference.

Score how well the model's review covers the points the human reviewers explicitly raised.

Human Reviews
{human\_reviews\_text}

Model's Review
{model\_review}

Issue Overlap
Extract every substantive point from the human reviews (strengths, weaknesses, questions). Check if the model's review covers each point (by substance, not exact wording).

CRITICAL: Only list points the human reviewers explicitly stated. Do NOT add points from your own assessment or knowledge. Every point in your list must be directly traceable to the human review text above.

Convergent points (raised by ≥ 2 reviewers) weight 2× single-reviewer points. Score = (weighted covered) / (total weighted).

Also extract:

reference\_verdict.overall\_mean: mean of human rating scores
reference\_verdict.decision\_consensus: "Accept" if >50\% vote Accept, else "Reject"
reference\_verdict.soundness\_mean: mean soundness if reported, else null
Return JSON. No preamble.

\end{promptbox}

\begin{promptbox}{Fabrication}
    Call 2 --- Fabrication Inputs: full paper body + AI review (no human reviews)

You are an experienced area chair at a top-tier ML conference.

Check whether the model's review fabricates facts about the paper.

Full Paper Body (up to References)

\{paper\_body\}

Model's Review

\{model\_review\}

Fabrication Check

Enumerate every specific factual claim in the model review: numbers, named baselines, algorithm details, section references, quoted phrases.

For each claim, scan the paper body above to verify it. The paper is the only source of truth.

Numbers may appear as prose (0.5), LaTeX (\textbackslash\textbackslash( 0.5 \textbackslash\textbackslash)), or HTML table cells ($<$td$>$14.71\%$<$/td$>$). Names may appear in running text, reference lists, or abbreviations.

Mark each claim:

verified --- found in paper body. Quote matching text in note.

unverified --- paper positively contradicts the claim: a different value is explicitly present, or the paper states the opposite. Quote the contradiction in note.

unverifiable --- paper body neither confirms nor contradicts (e.g. claim is about a result, number, or detail that would plausibly live in a table or figure absent from the converted text). No penalty.

Before marking unverified you must:

(1) check all text forms (prose, LaTeX math, HTML table cells),

(2) confirm an explicit contradiction exists --- not finding something is \textbf{NOT} a contradiction,

(3) ask whether the value could be in a missing table or figure --- if yes, mark unverifiable.

When in doubt between unverified and unverifiable, choose unverifiable.

Score: 1.0 if zero unverified, 0.5 if one, 0.0 if $\geq$ 2.

Return JSON. No preamble.
\end{promptbox}

\begin{promptbox}{Calibration}
    Call 3 --- Calibration

Inputs: paper title + abstract + full paper body + human reviews + AI review

You are an experienced area chair at a top-tier ML conference.

Score the model's review on: Comprehension, Substance, Insight, and Calibration.

Paper

**Title:** \{title\}

**Abstract:** \{abstract\}

**Full Paper Body (up to References):**

\{paper\_body\}

 Human Reviews

\{human\_reviews\_text\}

Model's Review

\{model\_review\}

---

Criterion 1: Pairwise Calibration --- continuous [0.0, 1.0]

Score two dimensions and take the mean:

**Dimension A --- Decision Alignment** (binary: 0 or 1):

- Extract the model's final verdict (Accept/Reject) from its review.
- Extract the human consensus verdict from the human reviews above.
- Score 1.0 if they match, 0.0 if they do not.

**Dimension B --- Internal Consistency** (continuous: 0.0, 0.5, 1.0):

- Read the model's Strengths, Weaknesses, and Overall score together.
- Score 1.0 if the verdict and score are fully consistent with the written critique: a paper praised with few weaknesses gets a high score and Accept; a paper heavily criticised gets a low score and Reject.
- Score 0.5 if there is a minor mismatch (e.g. mostly positive but slightly low score, or one unexplained inconsistency).
- Score 0.0 if there is a clear contradiction: the review raises severe weaknesses but accepts the paper, or praises it strongly but rejects it.

Final calibration score = mean(Dimension A, Dimension B).

Return JSON. No preamble.
\end{promptbox}

%% file: SupplementExtra.tex
\section{Example AI Review}
  \label{app:annotated}
  The following content is an example review produced by our AI Reviewer and AppliedScientist for the paper mentioned below.

  Original manuscript: Gorka Abad, Stjepan Picek, Lorenzo Cavallaro, and Aitor Urbieta, *Context is the Key: Backdoor Attacks for In-Context Learning with Vision Transformers* (2026), OpenReview: \url{https://openreview.net/forum?id=1h7JvTbIYJ}.

\begin{quotebox}{Summary}
\footnotesize
This paper is an extended version of the authors' earlier preprint on
in-context backdoor attacks against masked-image-modeling vision
transformers (Painter ViT-L and SegGPT ViT-L). The carried-over
contributions are task-specific and task-agnostic data-poisoning backdoors.
The genuinely new content is a measurement correction, a causal mechanism,
and a defense: (1) the paper shows the standard degradation metric
conflates the backdoor with the fact that the trigger occludes 10\% of the
input, since a clean model fed the trigger already loses 22--42\% PSNR; it
introduces trigger-controlled degradation and a targeted attack success
rate (TASR) that remove this confounder, and re-measures the attack with
them; (2) it attributes the attack to a ``context-routing hijack'' and
demonstrates causality by projecting out an estimated trigger direction at
inference; (3) it proposes context-consistency screening (CCS), a
black-box detector that separates triggered from clean inputs (AUC 1.00 on
the test set). The attack and its cross-task transfer are reproduced on a
second backbone (SegGPT).
\end{quotebox}

\begin{quotebox}{Strength --- self-correction}
\footnotesize
The occlusion-confounder analysis is the strongest part. The exact
within-construction decomposition --- clean/triggered = 22.76$\to$15.84,
compromised/triggered = 5.37, giving 76.4\% total = 30.4\% occlusion +
46.0\% backdoor --- is clearly stated and arithmetically consistent. The
paper explicitly corrects its own earlier evaluation (``our earlier
evaluation understated cross-task transfer because its metric could not
see it''), which is rare and commendable.
\end{quotebox}

\begin{quotebox}{Weakness --- test construction for the headline claim}
\footnotesize
The never-poisoned-task TASR values (deraining 0.608, denoising 0.902,
grayscale 0.914, against 0.152--0.170 clean) are measured on tasks built
from LoL images, not on the native SRD/SIDD pipelines. This matters because
the earlier native-dataset evaluation found the LoL attack produced
``small or almost no degradation'' on deraining/denoising. The direction
is consistent across both evaluations, but the magnitude is not directly
comparable, and the paper's most distinctive claim --- task-agnostic
behavior from a task-specific budget --- is currently supported only in a
construction where image statistics are held fixed.
\end{quotebox}

\begin{quotebox}{Weakness --- detection results are a 15-query result}
\footnotesize
CCS, STRIP, activation clustering, and spectral signatures all report
AUC/accuracy 1.00 on the compromised model, but on only 15 held-out LoL
test queries. An AUC computed over ordered pairs of 15 queries cannot
distinguish 1.00 from, say, 0.99, and the defense section is where the
weakest statistics support the strongest-sounding numbers. CCS is also
only validated on LoL, not on the native task suite or on SegGPT.
\end{quotebox}

\begin{quotebox}{Suggestion, tied to the weakness above}
\footnotesize
Run TASR on the native pipelines (SRD, SIDD, ADE20K, NYUv2, FSS-1000) for
the LoL-poisoned model. This is the single experiment that would convert
the cross-task transfer claim from ``demonstrated in a controlled
construction'' to ``demonstrated in the paper's own primary evaluation
protocol.'' Report TASR and trigger-controlled degradation alongside the
existing degradation values for the same native test sets so the two views
are directly comparable.
\end{quotebox}

\begin{quotebox}{Scores}
\footnotesize
\textbf{Soundness} 3/4 --- The confounder analysis, decomposition, causal
intervention, and specificity controls are exemplary; the cross-task-transfer
and detection claims rest on nonstandard or tiny test constructions.
\textbf{Presentation} 3/4 --- Dense but clear, and unusually honest about
limitations; marred by a handful of small numerical inconsistencies across
sections.
\textbf{Contribution} 3/4 --- The attack itself is prior work, but the
measurement correction, the context-routing mechanism, and the black-box
defense are meaningful, citable additions that improve the whole
sub-field's evaluation practice.
\textbf{Overall} 6/10. \textbf{Confidence} 4/5. \textbf{Decision}: Accept.
\end{quotebox}